%% file: valhalla.tex
\documentclass{gamedojo}
\usepackage{colortbl}
\usepackage{pifont}
\usepackage{adjustbox}
\usepackage{fontawesome5}
\usepackage{placeins}
\newcommand{\cmark}{\textcolor{green!55!black}{\ding{51}}}
\newcommand{\xmark}{\textcolor{red!70!black}{\ding{55}}}
\newcommand{\pmark}{\textcolor{orange!90!black}{\ding{119}}}
\newcommand{\gamedojo}{\textsc{GameDojo}}
\newcommand{\gamedojoMainRef}[2]{\ref{#1}}

\newif\ifgamedojoUsageStatement
\gamedojoUsageStatementfalse
\hypersetup{
  pdftitle={GameDojo: Benchmarking Real-Time Interactive Agents in Video Games},
  pdfauthor={Kuan Zhang, Zhihao Yang, Yukun Chen, Xingyiliang Yin, Mingyu Ouyang, Yue Su, Xiangnan Wu, Jiaming Li, Longze Chen, Run Luo, Zirong Chen, Lu Wang, Yiming Li},
  pdfsubject={Interactive intelligence, generalist agents, and controllable boss encounters}
}
\title{\textcolor{GameInk}{Game}\textcolor{GameLink}{Dojo}: Benchmarking Real-Time\\Interactive Agents in Video Games}
\newcommand{\gamedojoAuthorNames}{%
  Kuan Zhang\textsuperscript{1,*},\enspace Zhihao Yang\textsuperscript{2,*},\enspace
  Yukun Chen\textsuperscript{2,*},\enspace Xingyiliang Yin\textsuperscript{3},\enspace
  Mingyu Ouyang\textsuperscript{4},\enspace Yue Su\textsuperscript{5},\\
  Xiangnan Wu\textsuperscript{2},\enspace Jiaming Li\textsuperscript{2},\enspace
  Longze Chen\textsuperscript{2},\enspace Run Luo\textsuperscript{4},\enspace
  Zirong Chen\textsuperscript{1},\enspace Lu Wang\textsuperscript{1,\textdagger},\enspace
  Yiming Li\textsuperscript{1,\textdagger}}
\newcommand{\gamedojoAffiliations}{%
  \textsuperscript{1}Tsinghua University\qquad
  \textsuperscript{2}University of the Chinese Academy of Sciences\\
  \textsuperscript{3}University of Macau\qquad
  \textsuperscript{4}National University of Singapore\qquad
  \textsuperscript{5}University of Hong Kong}
\newcommand{\gamedojoAuthorBlock}{\centering
  {\rmfamily\bfseries\fontsize{9.35}{12}\selectfont\gamedojoAuthorNames\par}
  \vspace{1.5mm}
  {\rmfamily\bfseries\itshape\fontsize{8.2}{10.5}\selectfont\color{GameSlate}\gamedojoAffiliations\par}
  \vspace{1.5mm}
  {\rmfamily\fontsize{8.2}{10.5}\selectfont\color{GameSlate}
    \textsuperscript{*}Equal contribution.\qquad
    \textsuperscript{\textdagger}Co-corresponding authors.\par}}

\author{\gamedojoAuthorBlock}

\newcommand{\gamedojoHuggingFaceIcon}{%
  \raisebox{-.18em}{\includegraphics[height=1.1em]{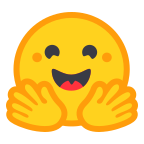}}}
\newcommand{\gamedojoResourceItem}[3]{%
  \mbox{\href{#3}{\textcolor{GameInk}{#1}\enspace\textcolor{GameLink}{#2}}}}
\newcommand{\gamedojoResourceBlock}{%
  \par\vspace{4mm}
  {\centering\rmfamily\small\bfseries
    \gamedojoResourceItem{\faGlobe}{Project Page}{https://gamedojo-benchmark.github.io/}\hspace{5mm}%
    \gamedojoResourceItem{\faGithub}{GitHub}{https://github.com/THUSI-Lab/GameDojo}\hspace{5mm}%
    \gamedojoResourceItem{\gamedojoHuggingFaceIcon}{Benchmark \& Dataset}{https://huggingface.co/datasets/THUSI-Lab/GameDojo}\hspace{5mm}%
    \gamedojoResourceItem{\gamedojoHuggingFaceIcon}{Model}{https://huggingface.co/THUSI-Lab/GameDojo-4B}\par}}
\newpagestyle{gamedojofirst}{%
  \sethead{}{}{}
  \setfoot{\scriptsize\gameaititlefont\color{GameSlate}\faEnvelope\enspace Email:
    \href{mailto:zhang-k26@mails.tsinghua.edu.cn}{zhang-k26@mails.tsinghua.edu.cn};\enspace
    \href{mailto:yimingli9702@mail.tsinghua.edu.cn}{yimingli9702@mail.tsinghua.edu.cn}}{}{}
  \setheadrule{0pt}\setfootrule{0pt}}

\newcommand{\gamedojoInstitutionLogo}[1]{\raisebox{-.5\height}{#1}}
\newcommand{\gameaiTitleMark}{%
  \noindent\makebox[\linewidth][c]{%
    \begin{minipage}{\dimexpr\linewidth+6mm\relax}
      \noindent\begin{minipage}[c]{.35\linewidth}
        \raggedright
        \includegraphics[width=\linewidth,trim=4bp 0 0 0,clip]{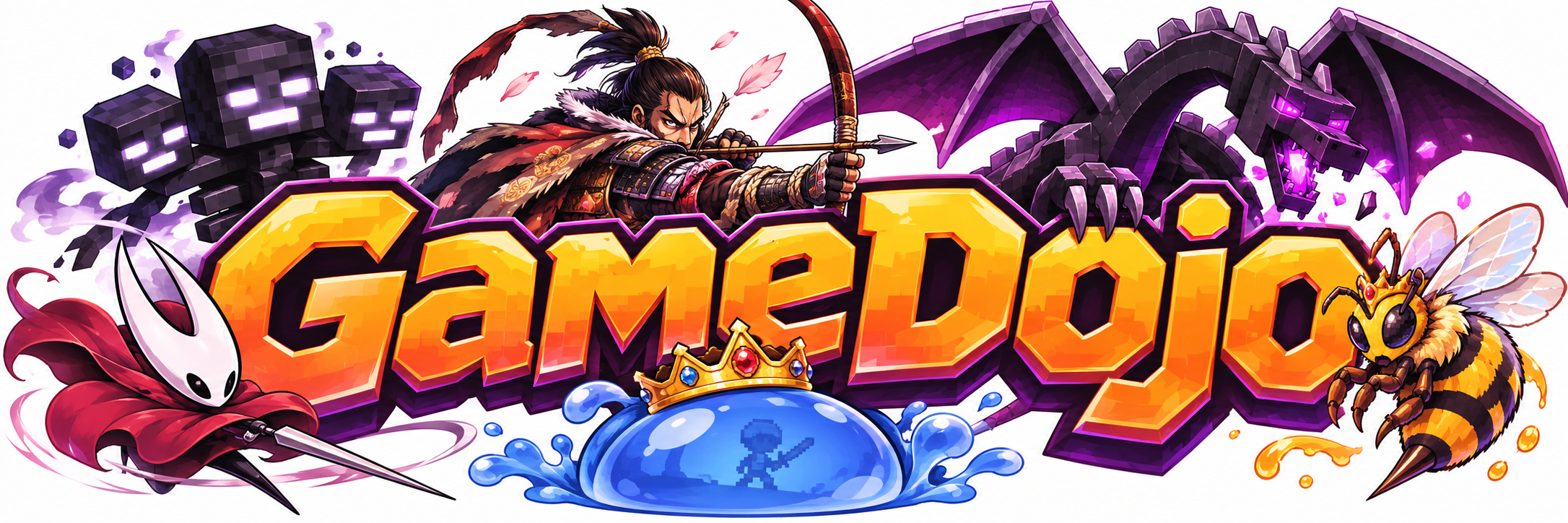}
      \end{minipage}\hfill%
      \begin{minipage}[c]{.625\linewidth}
        \noindent\makebox[\linewidth]{%
          \gamedojoInstitutionLogo{\includegraphics[height=9.3mm,trim=229bp 143bp 276bp 152bp,clip]{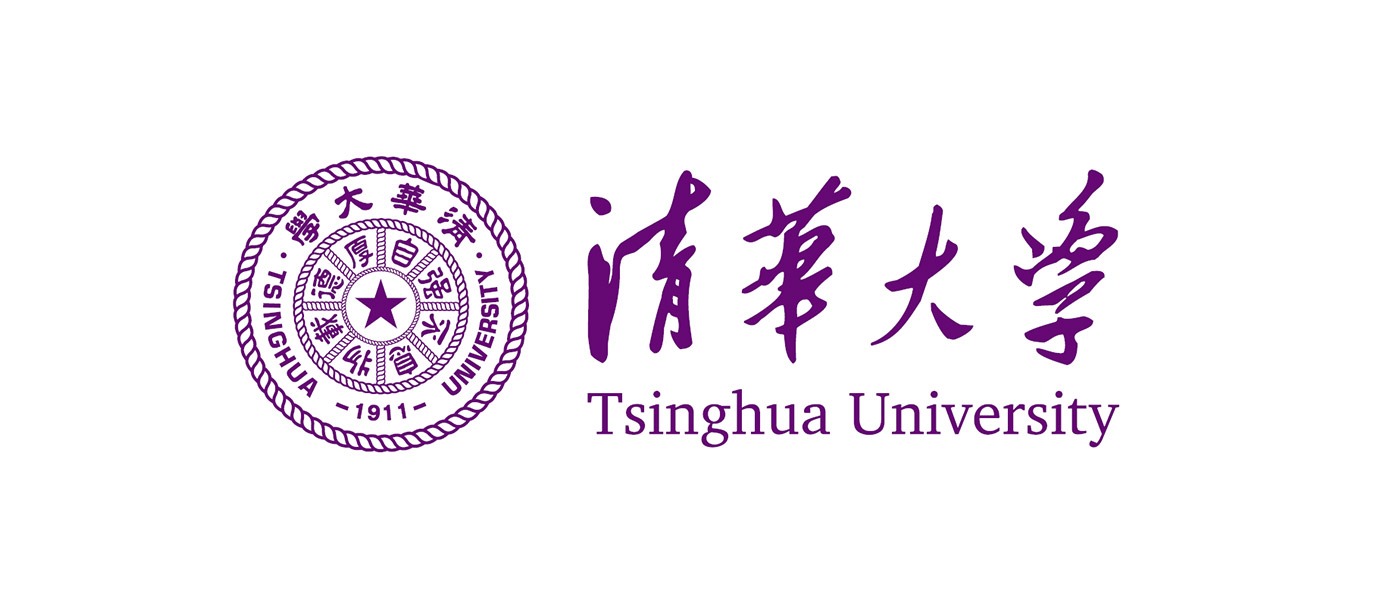}}\hfill
          \gamedojoInstitutionLogo{\includegraphics[height=9.4mm,trim=49.2bp 50.64bp 49.2bp 47.52bp,clip]{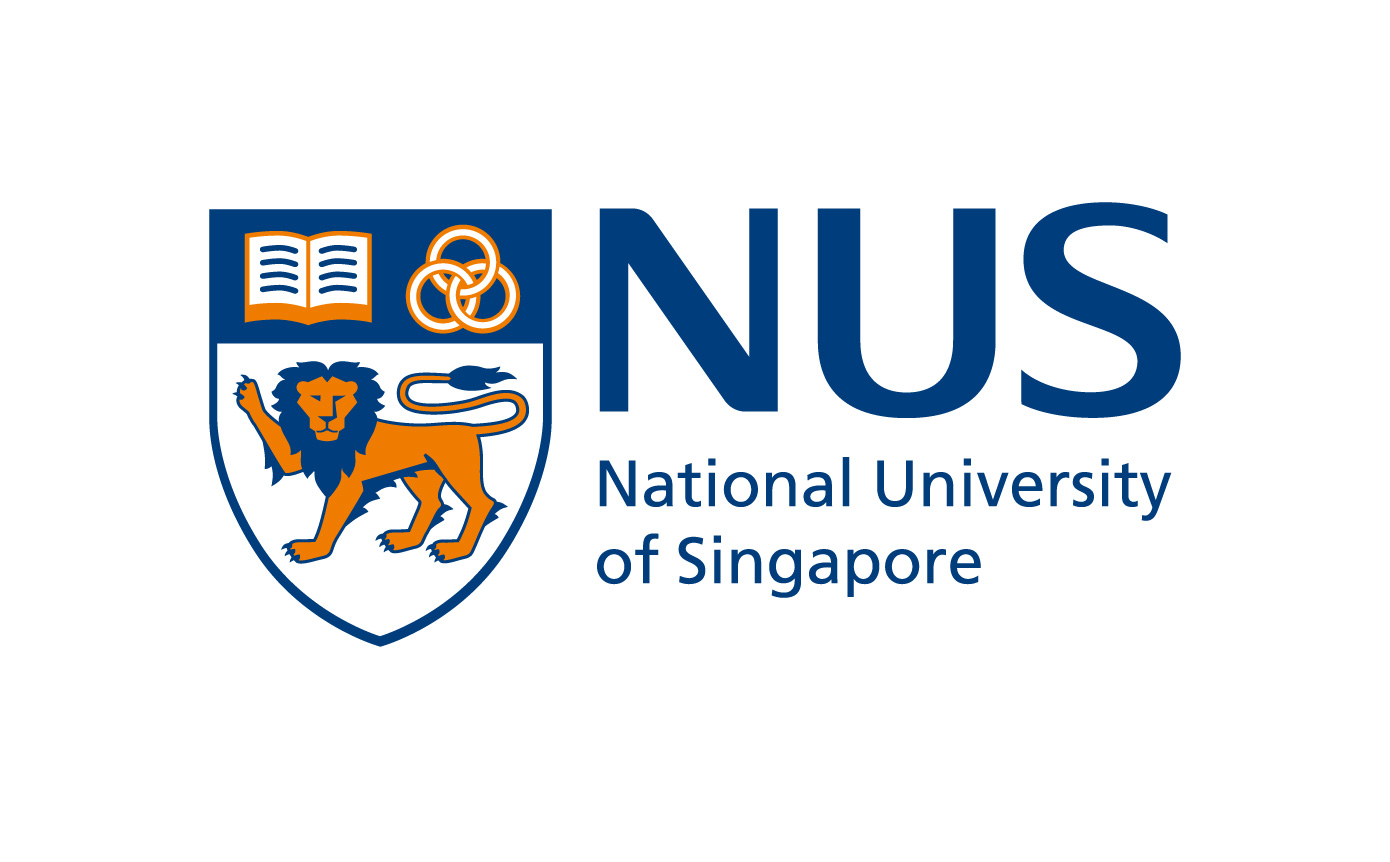}}\hfill
          \gamedojoInstitutionLogo{\includegraphics[height=10.1mm,trim=0 0 0 0,clip]{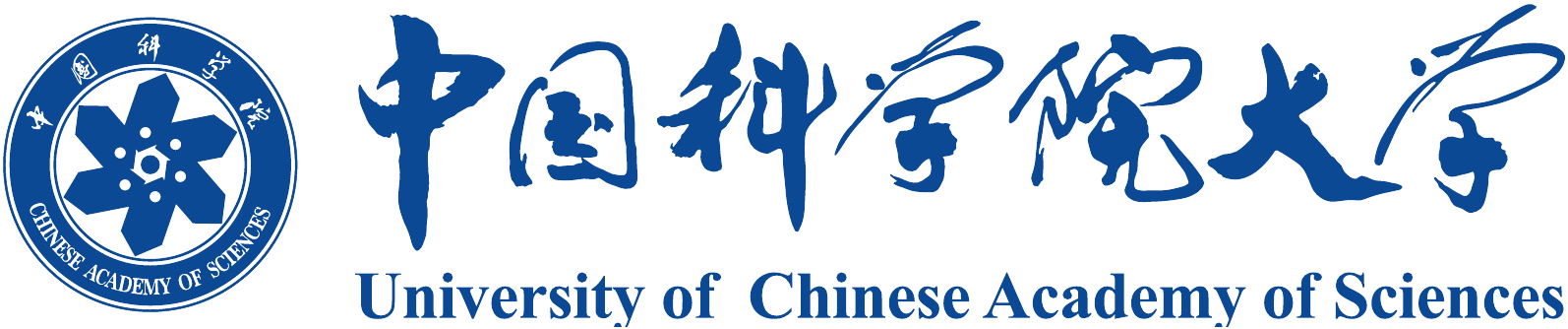}}}\par
        \vspace{2mm}
        \noindent\makebox[\linewidth][r]{%
          \gamedojoInstitutionLogo{\includegraphics[height=10.2mm,trim=0 0 0 0,clip]{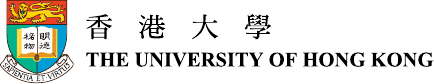}}%
          \hspace{5mm}%
          \gamedojoInstitutionLogo{\includegraphics[height=10.4mm,trim=43.49bp 153.73bp 26.25bp 149.23bp,clip]{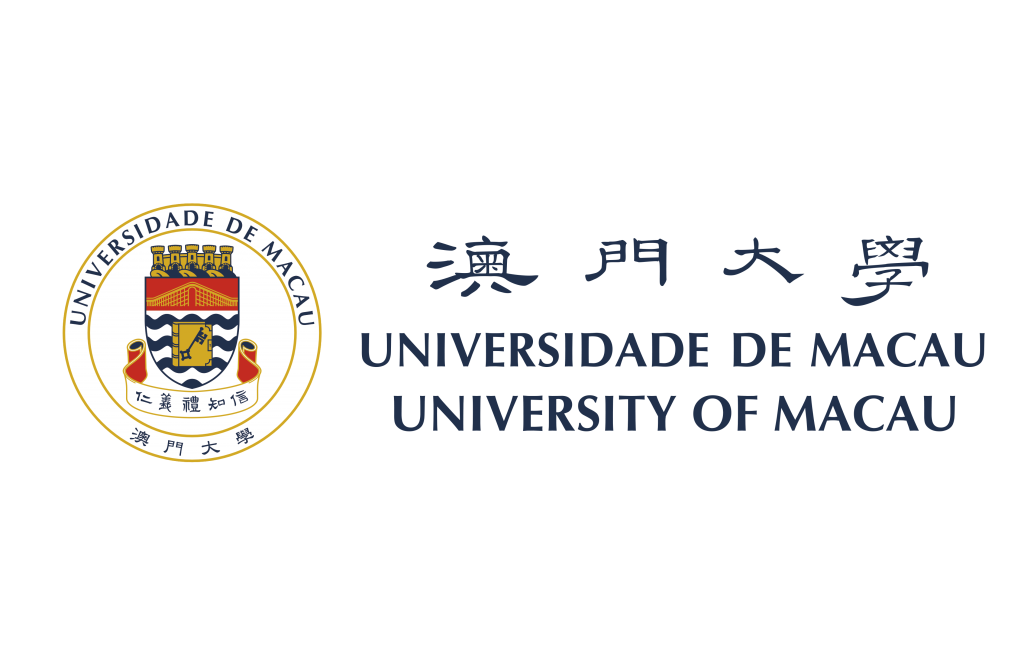}}}\par
      \end{minipage}
    \end{minipage}}\par
  \vspace{4mm}
  {\color{black}\hrule height .5pt}
  \vspace{4mm}}
\renewenvironment{gameaifrontmatter}{%
  \thispagestyle{gamedojofirst}\noindent\begin{minipage}{\linewidth}
  \setlength{\parindent}{0pt}\setlength{\parskip}{0pt}
  \gameaiTitleCore\vspace{5mm}
}{\end{minipage}\par\vspace{2mm}}
\input{sections/experiment_floats}
\begin{document}
\begin{gameaifrontmatter}
  {\small\color{GameInk}\justifying\noindent
Interactive intelligence requires agents to act on observations and adapt to the consequences of actions.
Studying these abilities calls for diverse, challenging tasks under controlled conditions. We introduce \gamedojo{}, a controllable interactive gameplay benchmark of 120 boss encounters across 20 games for evaluating model capabilities, learning algorithms, and agent harnesses in interaction. We have the coding agents reference classic games and scenes to build the environment and write gameplay scripts to collect gameplay data automatically, producing \gamedojo{}-Data with 600 hours of high-quality gameplay trajectories. Experiments reveal substantial challenges for frontier models, especially in real-time interaction where all models win 0 score. Training on \gamedojo{}-Data brings \gamedojo{}-4B to frontier performance, with the strongest real-time performance. We also find joint training improves individual tasks, and learned policies transfer across bosses. Together, these findings establish \gamedojo{} as a challenging and valuable interactive gameplay benchmark for the community.
  \par}
  \gamedojoResourceBlock
\end{gameaifrontmatter}
\suppressfloats[t]
\begin{figure}[t]
  \centering
  \includegraphics[width=\linewidth,trim=0 0 0 24bp,clip]{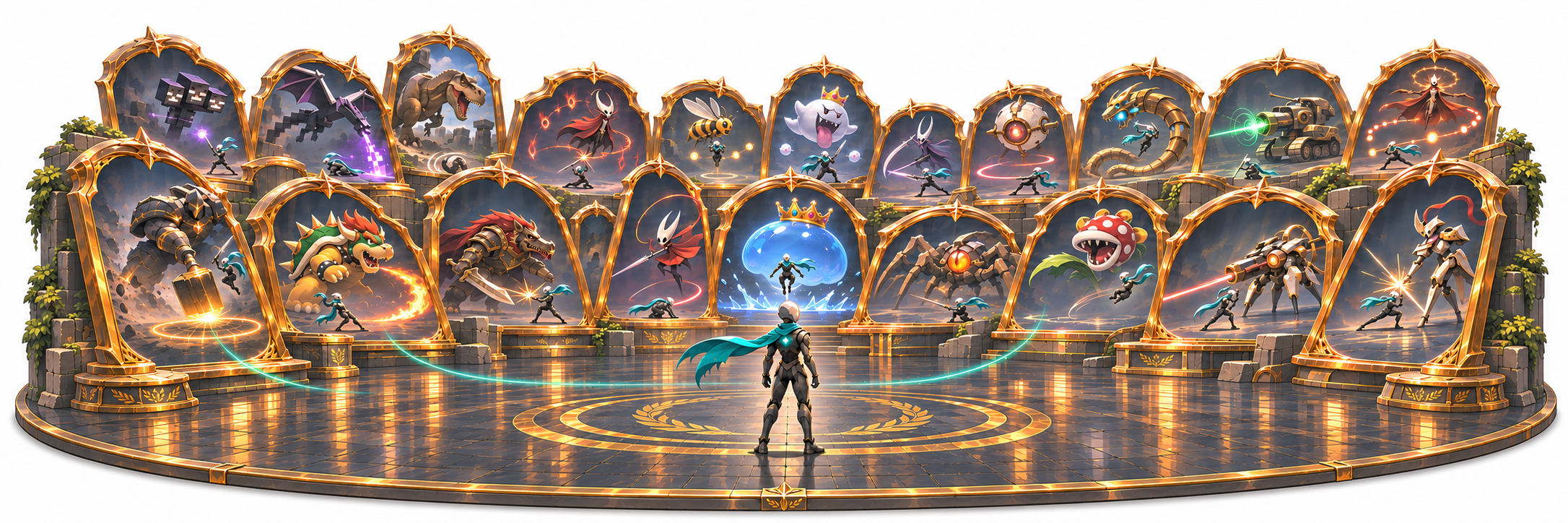}\par
  \captionsetup{skip=4pt}
  \caption{\textbf{\gamedojo{}: a controlled and challenging arena for interactive agents.}
    The benchmark spans 20 games and 120 boss encounters in diverse interaction patterns, supporting the study of model capabilities, algorithms, and agent harnesses.}
  \label{fig:teaser}
\end{figure}

\input{sections/01_introduction}

\gamedojoBenchmarkTable
\input{sections/02_related_work}
\gamedojoFrameworkFigure

\input{sections/03_environments}

\gamedojoTaskCoverageTable
\gamedojoFrontierTable

\input{sections/04_evaluation}

\gamedojoPpoFigure
\input{sections/05_experiments}
\gamedojoCrossGameFigure
\Needspace{14\baselineskip}
\input{sections/06_discussion}

\FloatBarrier
\setlength{\bibsep}{.6pt plus .2pt minus .2pt}
\bibliographystyle{plainnat}
\bibliography{references}

\clearpage
\appendix
\pdfbookmark[0]{Appendix}{gamedojo-appendix}
\section*{Appendix}
\suppressfloats[t]
\setcounter{table}{0}
\renewcommand{\thetable}{S\arabic{table}}
\renewcommand*{\theHtable}{appendix.\arabic{table}}
\input{sections/07_appendix}
\end{document}

%% file: sections/experiment_floats.tex
\newcommand{\gamedojoFrameworkFigure}{%
\begin{figure}[!t]
\centering
\includegraphics[width=\textwidth]{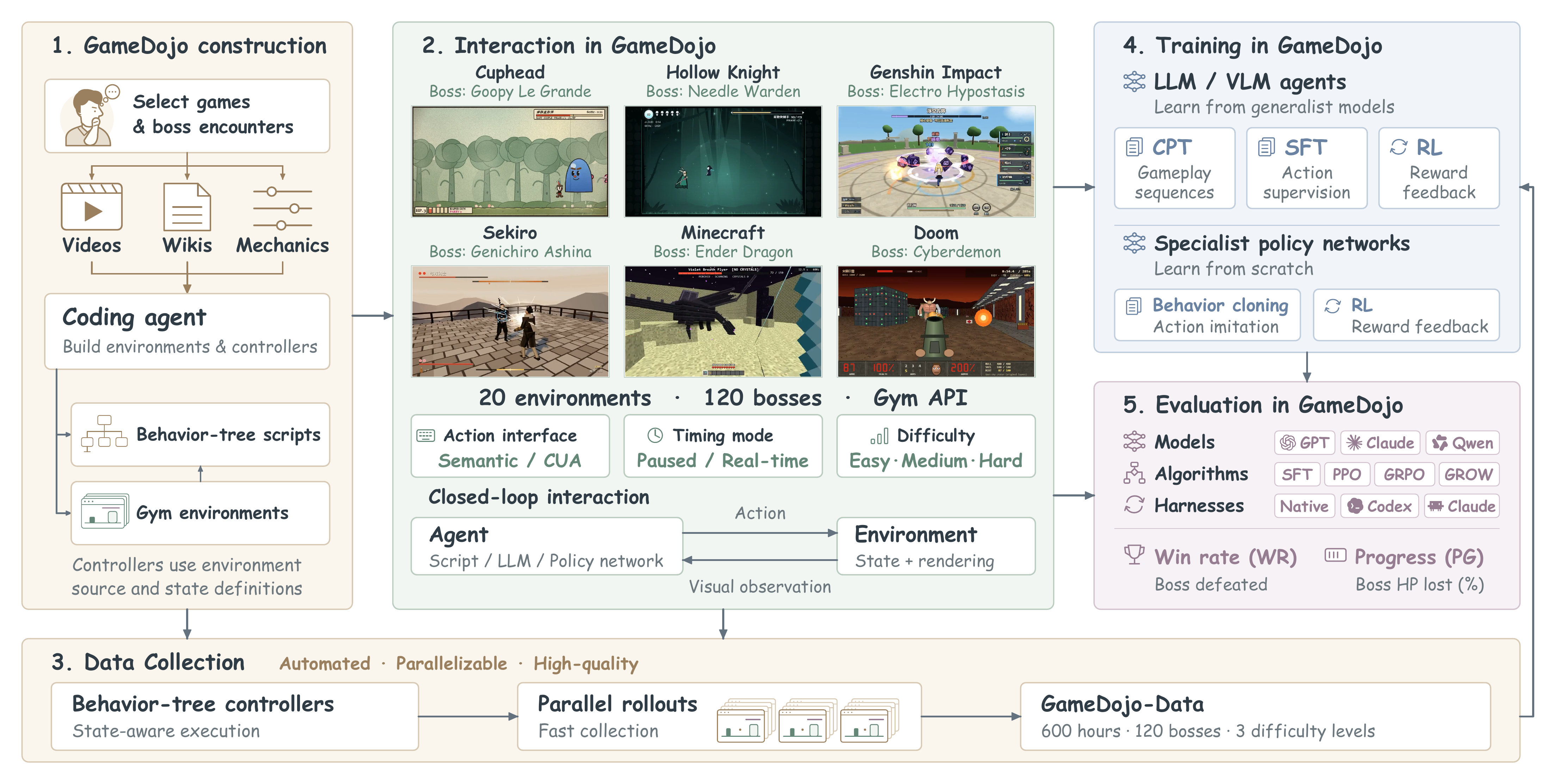}
\caption{\textbf{Overview of \gamedojo{}.}
(1) Human-selected games and bosses guide coding agents that build environments and behavior-tree controllers from videos, wikis, and mechanics descriptions.
(2) Scripts, LLM agents, and policy networks interact through configurable action interfaces, timing modes, and difficulty settings.
(3) Parallel rollouts produce high-quality \gamedojo{}-Data efficiently without any manual gameplay.
(4) The suite supports learning from generalist models and training specialist policies from scratch.
(5) Shared tasks evaluate model capabilities, algorithms, and agent harnesses.}
\label{fig:framework}
\end{figure}
}

\newcommand{\gamedojoBenchmarkTable}{%
\begin{table}[!t]
  \centering
  \caption{Comparison of related benchmarks. Games counts supported environments.
  Tasks counts benchmark objectives or boss encounters; Config. denotes configurable scenario sets.
  For game source, \cmark{} denotes fully available source and state access, \pmark{} partial availability, and \xmark{} closed-source games.
  Modern AAA indicates whether the benchmark includes AAA and modern commercial titles or tasks based on them; \gamedojo{} uses independently reconstructed encounters.
  Core capabilities summarizes task demands and interaction coverage. }
  \label{tab:benchmark-comparison}
  \footnotesize
  \setlength{\tabcolsep}{3pt}
  \begin{adjustbox}{width=\linewidth}
  \begin{tabular}{@{}lccccl@{}}
    \toprule
    Benchmark & Games & Tasks & Game source & Modern AAA & Core capabilities \\
    \midrule
    \rowcolor{black!10}
    \multicolumn{6}{l}{\textit{General game benchmarks}} \\
    VideoGameBench~\citep{zhang2025videogamebench} & 23 & 23 & \xmark{} & \xmark{} & Navigation, puzzles, combat, memory \\
    Orak~\citep{park2025orak} & 12 & 12 & \pmark{} & \cmark{} & Navigation, puzzles, combat, memory \\
    GameVerse~\citep{zhang2026gameverse} & 15 & 15 & \pmark{} & \cmark{} & Navigation, puzzles, combat, memory \\
    OmniGameArena~\citep{omnigamearena2026} & 12 & 12 & \cmark{} & \pmark{} & Navigation, combat, cooperation \\
    GameWorld~\citep{ouyang2026gameworld} & 34 & 170 & \cmark{} & \xmark{} & Navigation, puzzles, combat, memory \\
    PlayerBench~\citep{game2player2026} & 50 & 287 & \pmark{} & \xmark{} & Navigation, puzzles, combat, memory \\
    \midrule
    \rowcolor{black!10}
    \multicolumn{6}{l}{\textit{Adversarial environments}} \\
    Gym-$\mu$RTS~\citep{huang2021gymmicrorts} & 1 & Config. & \cmark{} & \xmark{} & 1: real-time strategy \\
    ViZDoom~\citep{kempka2016vizdoom} & 1 & Config. & \cmark{} & \xmark{} & 1: first-person shooting \\
    EvoMan~\citep{defranca2019evoman} & 1 & 8 & \cmark{} & \xmark{} & 1: action platforming \\
    SoulsGym~\citep{soulsGym} & 1 & 1 & \xmark{} & \cmark{} & 1: action adventure and role-playing \\
    DSLE~\citep{gezgin2026dsle} & 1 & 22 & \xmark{} & \cmark{} & 1: action adventure and role-playing \\
    \gamedojo{} (Ours) & \textbf{20} & \textbf{120} & \cmark{} & \cmark{} & \textbf{9 interaction patterns} (Table~\ref{tab:task-diversity}) \\
    \bottomrule
  \end{tabular}
  \end{adjustbox}
\end{table}
}

\newcommand{\gamedojoTaskCoverageTable}{%
\begin{table}[!t]
  \centering
  \caption{Task coverage: nine interaction patterns and reference games used in environment design.
  The grid shows one boss encounter from each of the 20 reconstructed environments.}
  \label{tab:task-diversity}
  \small
  \begin{adjustbox}{width=\linewidth}
  \begin{tabular}{@{}ll@{}}
    \toprule
    Interaction pattern & Reference games \\
    \midrule
    Action platforming & Hollow Knight; Kirby; Mega Man; Super Mario 64 \\
    Run-and-gun combat & Cuphead \\
    Fighting & The King of Fighters '97 \\
    Top-down combat & Battle City; Soul Knight; The Binding of Isaac \\
    Bullet-hell dodging & Touhou: Embodiment of Scarlet Devil \\
    First-person shooting & Doom; Quake II; Hexen II \\
    Action adventure and role-playing & Sekiro; Genshin Impact; Tomb Raider I; The Legend of Zelda: Ocarina of Time \\
    Sandbox survival combat & Minecraft; Terraria \\
    Racing & Racing environment \\
    \bottomrule
  \end{tabular}
  \end{adjustbox}\par\medskip
  \includegraphics[width=\linewidth]{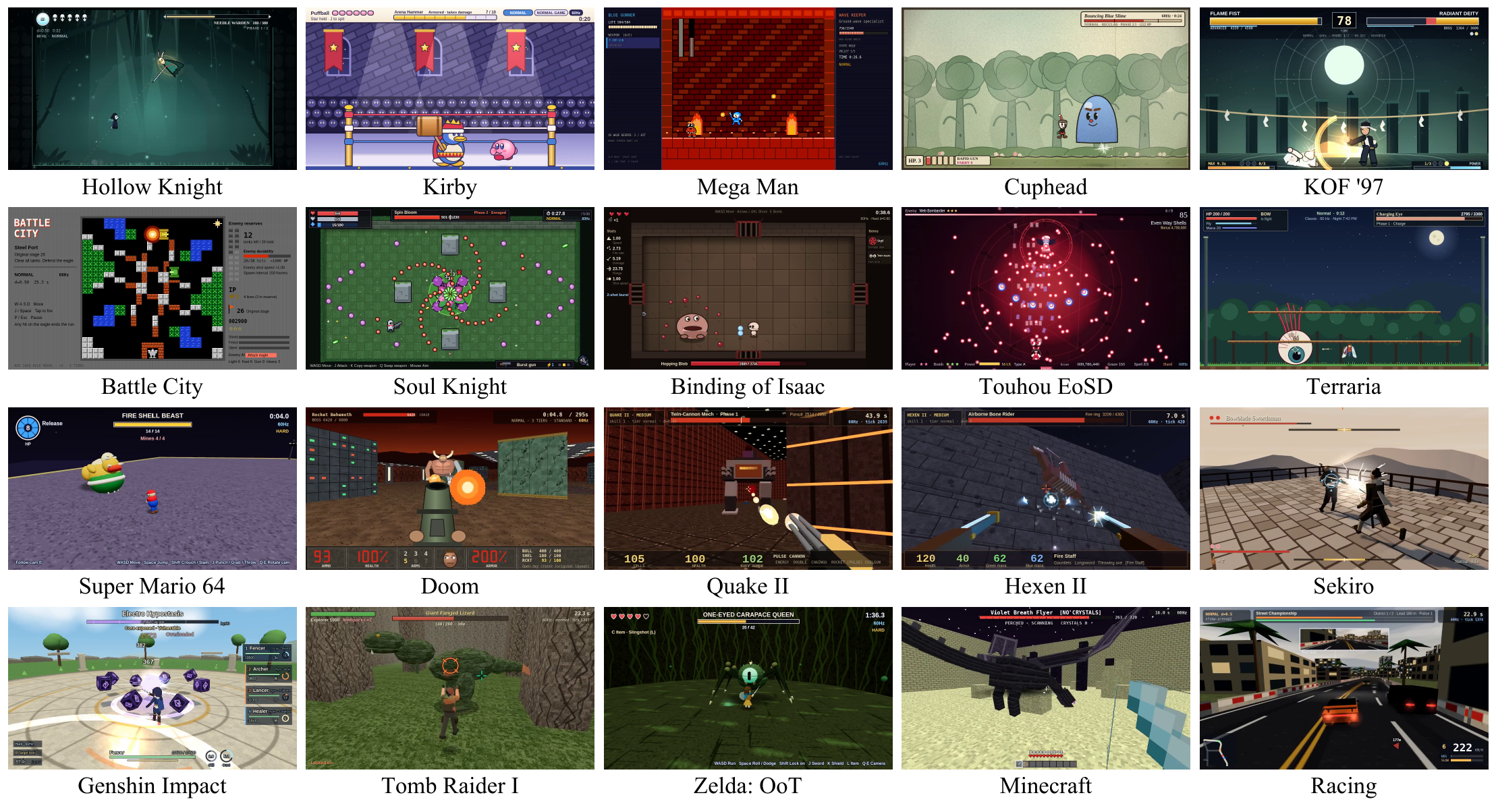}
\end{table}
}

\newcommand{\gamedojoFrontierTable}{%
\begin{table}[!t]
\centering
\small
\setlength{\tabcolsep}{3pt}
\caption{Frontier-model and baseline performance at medium difficulty. Entries report WR / PG (\%) over 120 bosses, averaging three episodes per boss for models and five for the script. Average is the mean across Semantic and CUA. The best and second-best model results per metric and column are in \textbf{bold} and \underline{underlined}.}
\label{tab:frontier-main}
\begin{adjustbox}{width=\linewidth}
\begin{tabular}{@{}lcccccc@{}}
\toprule
Medium Difficulty & \multicolumn{3}{c}{Time-paused (300 steps, 1 gameplay min)} & \multicolumn{3}{c}{Real-time (3 physical min)} \\
\cmidrule(lr){2-4}\cmidrule(lr){5-7}
Model & Semantic & CUA & Average & Semantic & CUA & Average \\
\midrule
\rowcolor{black!10}
\multicolumn{7}{l}{\textit{Baselines}} \\
Idle & 0.0 / 0.3 & 0.0 / 0.3 & 0.0 / 0.3 & 0.0 / 0.3 & 0.0 / 0.3 & 0.0 / 0.3 \\
Random & 0.1 / 8.5 & 0.0 / 6.8 & 0.1 / 7.7 & 0.0 / 8.6 & 0.0 / 6.8 & 0.0 / 7.7 \\
Boss-specific script & 19.8 / 86.5 & 29.0 / 90.3 & 24.4 / 88.4 & 99.2 / 99.7 & 99.8 / 100.0 & 99.5 / 99.9 \\
\midrule
\rowcolor{black!10}
\multicolumn{7}{l}{\textit{Closed-source models}} \\
GPT-6.1-Sol & \textbf{5.8} / \textbf{45.0} & \textbf{3.3} / \textbf{44.6} & \textbf{4.6} / \textbf{44.8} & \textbf{0.0} / \textbf{2.5} & \textbf{0.0} / \textbf{2.9} & \textbf{0.0} / \textbf{2.7} \\
GPT-6-Luna & 0.0 / 23.1 & 0.0 / 28.1 & 0.0 / 25.6 & 0.0 / \underline{1.8} & 0.0 / 1.5 & 0.0 / 1.6 \\
Claude-Opus-5.5 & 0.0 / \underline{32.0} & 0.0 / \underline{30.4} & 0.0 / \underline{31.2} & 0.0 / 1.2 & 0.0 / 1.2 & 0.0 / 1.2 \\
Claude-Haiku-5.5 & 0.0 / 16.3 & 0.0 / 15.1 & 0.0 / 15.7 & 0.0 / 1.2 & 0.0 / 1.3 & 0.0 / 1.3 \\
Claude-Haiku-4.5 & 0.0 / 9.6 & 0.0 / 9.0 & 0.0 / 9.3 & 0.0 / 0.8 & 0.0 / 0.6 & 0.0 / 0.7 \\
Gemini-3.8-Flash & 0.0 / 31.1 & \underline{0.8} / 21.0 & \underline{0.4} / 26.1 & 0.0 / \underline{1.8} & 0.0 / 1.8 & 0.0 / 1.8 \\
\midrule
\rowcolor{black!10}
\multicolumn{7}{l}{\textit{Open-source models}} \\
DeepSeek-V4.1-Flash & 0.0 / 20.6 & 0.0 / 20.6 & 0.0 / 20.6 & 0.0 / 1.0 & 0.0 / 1.0 & 0.0 / 1.0 \\
GLM-5.3-Flash & 0.0 / 18.9 & 0.0 / 23.4 & 0.0 / 21.2 & 0.0 / 0.4 & 0.0 / 0.5 & 0.0 / 0.5 \\
MiniMax-M3 & 0.0 / 24.3 & 0.0 / 25.4 & 0.0 / 24.8 & 0.0 / 0.5 & 0.0 / 0.5 & 0.0 / 0.5 \\
Qwen3.5-35B-A3B & 0.0 / 12.7 & 0.0 / 10.3 & 0.0 / 11.5 & 0.0 / 0.6 & 0.0 / 0.5 & 0.0 / 0.6 \\
Qwen3.5-4B & 0.0 / 7.7 & 0.0 / 1.9 & 0.0 / 4.8 & 0.0 / 0.7 & 0.0 / 1.2 & 0.0 / 1.0 \\
\midrule
\rowcolor{black!10}
\multicolumn{7}{l}{\textit{Generalist game model}} \\
GameScaling-9B-S3 & - & 0.0 / 5.3 & 0.0 / 5.3 & - & 0.0 / 1.5 & 0.0 / 1.5 \\
Gamer-35B-A3B & - & 0.0 / 3.3 & 0.0 / 3.3 & - & 0.0 / \underline{2.0} & 0.0 / \underline{2.0} \\
\bottomrule
\end{tabular}
\end{adjustbox}
\end{table}
}

\newcommand{\gamedojoPpoFigure}{%
\begin{figure}[!t]
\centering
\resizebox{\textwidth}{!}{%
  \includegraphics[trim=0bp 0bp 844.32bp 0bp,clip]{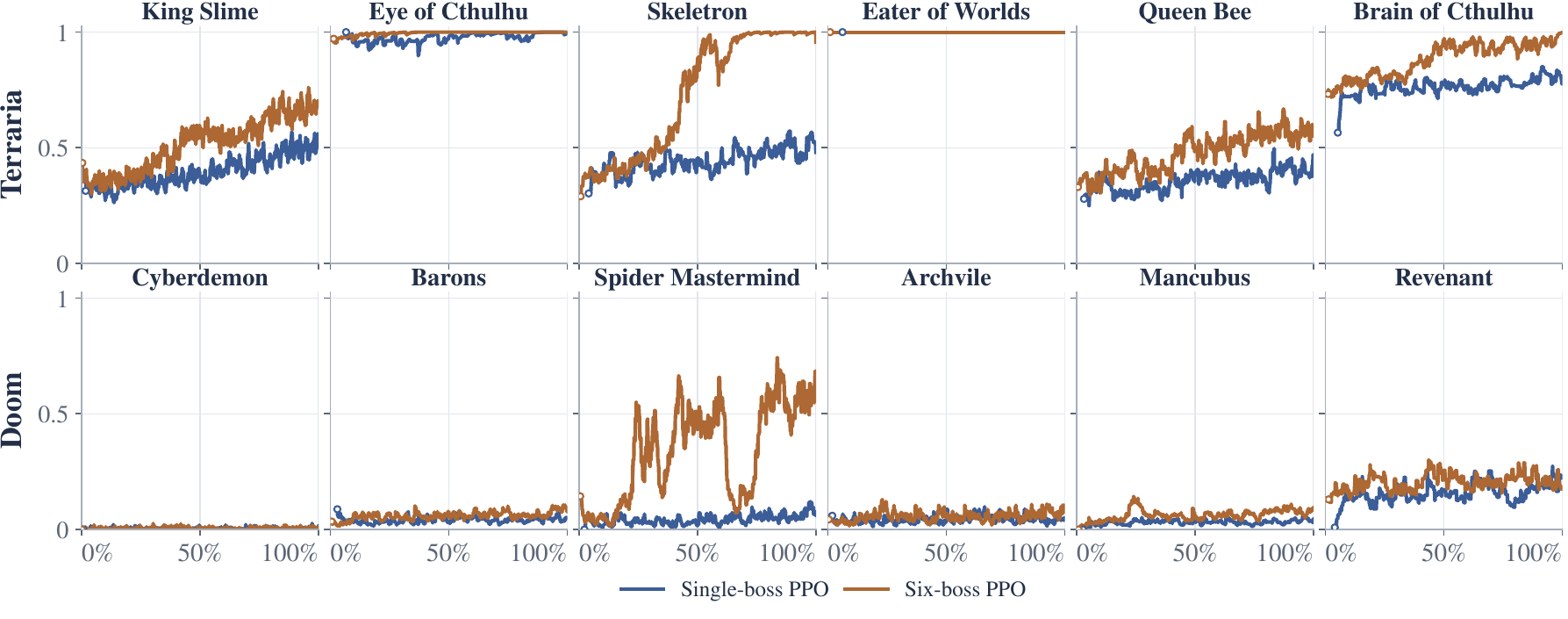}%
  \includegraphics[trim=19.2bp 0bp 0bp 0bp,clip]{assets/figures/ppo-boss-learning-curves.pdf}%
}
\caption{\textbf{PPO learning with single-boss and six-boss training.}
Terraria and Doom occupy the top and bottom rows, respectively.
Blue denotes a policy trained on the plotted boss, and orange denotes a policy trained jointly on all six bosses in the same game.
Both settings allocate 100K steps per boss (600K steps for the joint policy).
The axes show training progress and the fraction of boss health removed.}
\label{fig:ppo-boss-learning}
\end{figure}
}

\newcommand{\gamedojoBossCheckpointFigures}{%
\begin{figure}[!tp]
\centering
\includegraphics[width=\linewidth]{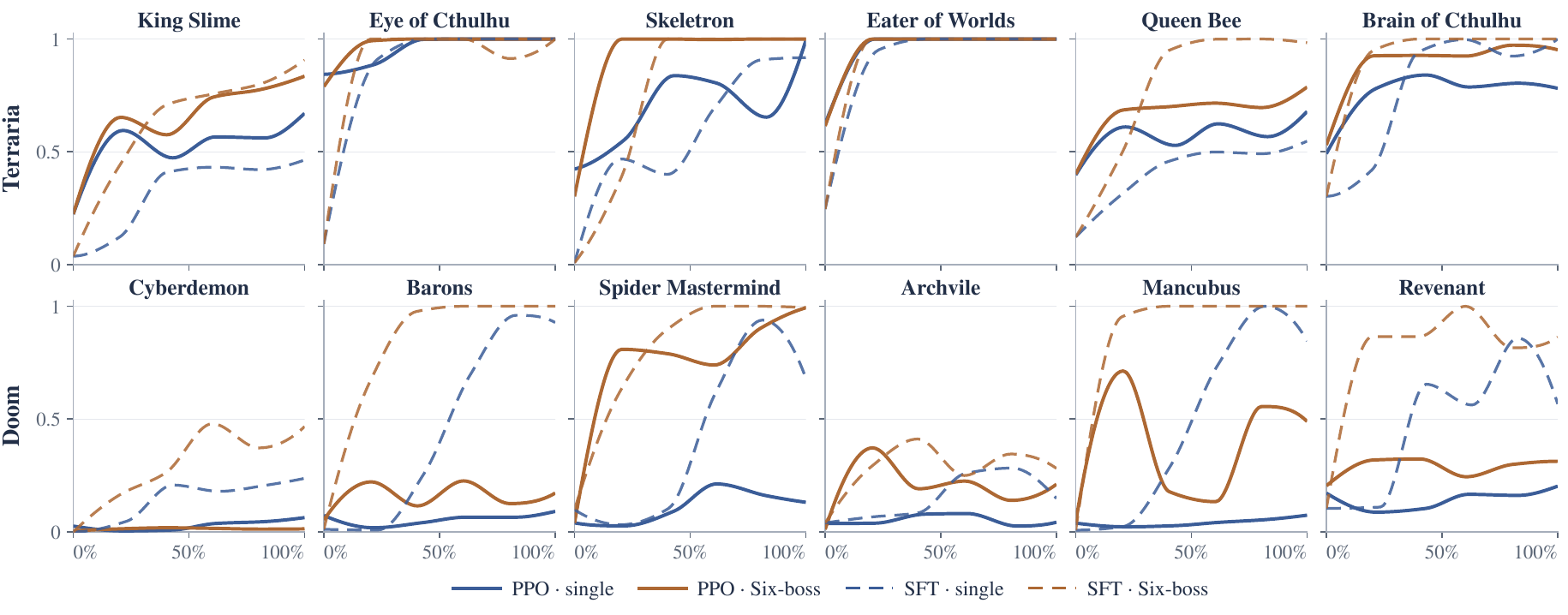}
\caption{\textbf{Effect of joint training on individual bosses.}
For each target boss in Terraria and Doom, we compare training on that boss alone (blue) with joint training on all six bosses in the same game (orange).
Solid curves show PPO evaluated with greedy argmax, and dashed curves show SFT.
Each curve interpolates six evaluations from 0\% to 100\% of training in 20\% increments.
The vertical axis shows the fraction of boss health removed.}
\label{fig:boss-joint-checkpoints}

\par\medskip
\captionsetup{type=figure}
\includegraphics[width=\linewidth]{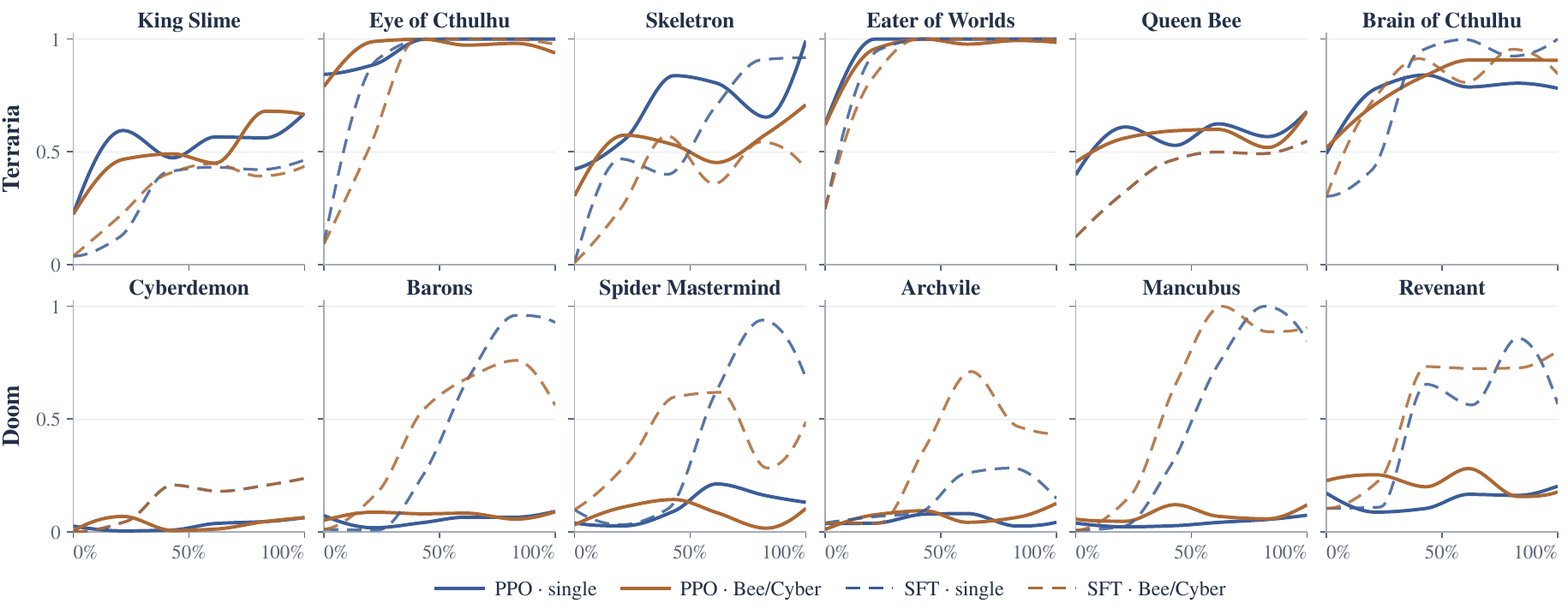}
\caption{\textbf{Transfer from a fixed training boss.}
For each target boss, we compare training on that boss (blue) with training on a fixed source (orange), Queen Bee in Terraria and Cyberdemon in Doom.
Solid curves show PPO evaluated with greedy argmax, and dashed curves show SFT.
The curves interpolate evaluations every 20\% of training from initialization to completion.
The fixed-source panels are shown for reference. The transfer averages cover only the five other bosses in each game.}
\label{fig:boss-transfer-checkpoints}
\end{figure}
}

\newcommand{\gamedojoCrossGameFigure}{%
\begin{figure}[!t]
\centering
\includegraphics[width=\textwidth]{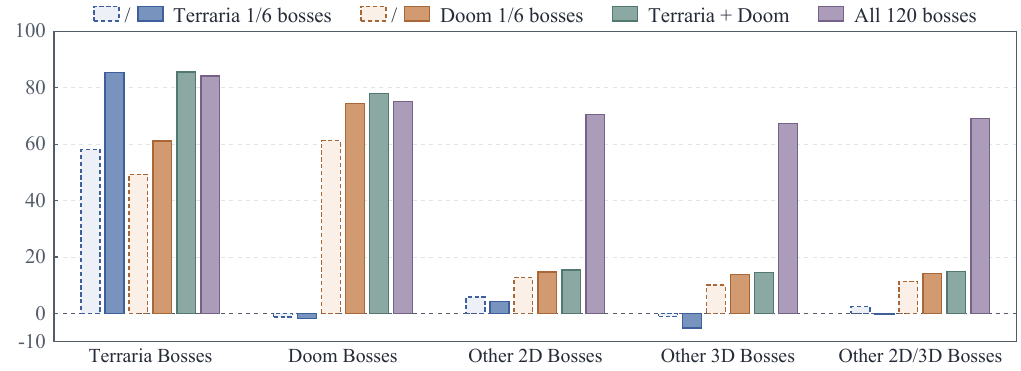}
\caption{\textbf{Evaluation on the performance of 6 models on 5 boss sets.} Mean progress gains (percentage points) over Qwen3.5-4B after SFT on \gamedojo{}-Data, evaluated with time-paused CUA on easy difficulty.
Training uses one or six bosses from Terraria or Doom, both games, or all 120 bosses (\gamedojo{}-4B); single-boss sources are Queen Bee and Cyberdemon, respectively.
The five target groups contain 6, 6, 54, 54, and 108 bosses; the last three exclude Terraria and Doom. All settings use the same 2-hour training data per boss.}
\label{fig:cross-game-transfer}
\end{figure}
}

\newcommand{\gamedojoDifficultyTable}{%
\begin{table}[!t]
\centering
\footnotesize
\setlength{\tabcolsep}{1.3pt}

\caption{Performance across difficulty levels using CUA, with a native-control script reference. Entries report WR / PG (\%) over 120 bosses; Average is the mean across difficulty levels. Models use three episodes per boss; the script uses five.}
\label{tab:difficulty}
\begin{adjustbox}{width=\linewidth}
\begin{tabular}{@{}lcccccccc@{}}
\toprule
CUA Mode & \multicolumn{4}{c}{Time-paused (300 steps, 1 gameplay min)} & \multicolumn{4}{c}{Real-time (3 physical min)} \\
\cmidrule(lr){2-5}\cmidrule(lr){6-9}
Model & Easy & Medium & Hard & Average & Easy & Medium & Hard & Average \\
\midrule
\rowcolor{black!10}
\multicolumn{9}{l}{\textit{Baselines}} \\
Random & 9.7 / 19.0 & 0.0 / 6.8 & 0.0 / 2.3 & 3.3 / 9.4 & 9.4 / 19.2 & 0.0 / 6.8 & 0.0 / 2.3 & 3.1 / 9.4 \\
Boss-specific script & 100.0 / 100.0 & 29.0 / 90.3 & 0.0 / 32.9 & 43.0 / 74.4 & 100.0 / 100.0 & 99.8 / 100.0 & 35.8 / 93.2 & 78.6 / 97.7 \\
\midrule
\rowcolor{black!10}
\multicolumn{9}{l}{\textit{Closed-source models}} \\
GPT-6.1-Sol & \textbf{58.3} / \textbf{78.1} & \underline{3.3} / \textbf{44.6} & \textbf{0.0} / \textbf{13.2} & \underline{20.6} / \textbf{45.3} & \underline{2.5} / \underline{13.7} & \textbf{0.0} / \underline{2.9} & \textbf{0.0} / \underline{0.7} & \underline{0.8} / \underline{5.8} \\
GPT-6-Luna & 25.8 / 58.6 & 0.0 / 28.1 & 0.0 / 8.2 & 8.6 / 31.6 & 0.0 / 7.1 & 0.0 / 1.5 & 0.0 / 0.5 & 0.0 / 3.0 \\
\midrule
\rowcolor{black!10}
\multicolumn{9}{l}{\textit{Open-source models}} \\
\gamedojo{}-4B & \underline{52.2} / \underline{76.7} & \textbf{11.9} / \underline{40.3} & \textbf{0.0} / \underline{11.8} & \textbf{21.4} / \underline{42.9} & \textbf{6.4} / \textbf{26.3} & \textbf{0.0} / \textbf{8.2} & \textbf{0.0} / \textbf{2.5} & \textbf{2.2} / \textbf{12.3} \\
Qwen3.5-4B & 0.3 / 6.7 & 0.0 / 1.9 & 0.0 / 0.6 & 0.1 / 3.0 & 0.0 / 2.8 & 0.0 / 1.2 & 0.0 / 0.4 & 0.0 / 1.5 \\
\midrule
\rowcolor{black!10}
\multicolumn{9}{l}{\textit{Generalist game models}} \\
GameScaling-9B-S3 & 4.2 / 16.2 & 0.0 / 5.3 & 0.0 / 1.8 & 1.4 / 7.8 & 0.0 / 5.8 & 0.0 / 1.5 & 0.0 / 0.5 & 0.0 / 2.6 \\
Gamer-35B-A3B & 4.2 / 12.9 & 0.0 / 3.3 & 0.0 / 1.0 & 1.4 / 5.7 & 0.8 / 6.6 & 0.0 / 2.0 & 0.0 / 0.5 & 0.3 / 3.1 \\
\bottomrule
\end{tabular}
\end{adjustbox}
\end{table}
}

%% file: sections/01_introduction.tex
\section{Introduction}
\label{sec:introduction}
Generalist agents must achieve goals through interaction with diverse environments~\citep{reed2022gato,simateam2024sima,simateam2025sima2,wang2025gametars,lumine2025, zhang2026gamescaling, luo2026gamer}.
This requires interactive intelligence: understanding the current situation, anticipating the consequences of actions,
and continually adjusting decisions based on feedback~\citep{yao2023react,shinn2023reflexion,wang2023voyager,zhang2026gamescaling, luo2026gamer}.
Adversarial tasks concentrate these demands. An opponent continually changes the situation and opportunities for action,
requiring agents to anticipate its behavior, choose a response, and act at the right moment through precise control~\citep{vinyals2019alphastar,defranca2019evoman,chen2025combatvla,wang2026roar, zhang2026gamescaling,luo2026gamer}.
These tasks provide a concrete setting for examining how perception, prediction, decision-making, and control
work together to produce effective behavior.

Building diverse, repeatable adversarial tasks in the physical world is costly and can pose safety risks,
making video games a safe and controllable environment~\citep{bellemare2013ale}.
However, whole-game evaluations combine navigation, puzzle solving, and combat~\citep{zhang2025videogamebench,park2025orak,zhang2026gameverse}, making it difficult to locate specific bottlenecks.
Open-source environments often feature older or small games~\citep{kempka2016vizdoom,cobbe2020procgen}, while modern games are typically closed-source~\citep{tan2024cradle,soulsGym}.
Boss encounters offer a focused task unit in which recognizing attack cues, timing actions, and adjusting strategies directly affect progress~\citep{chen2025combatvla,soulsGym}. Implementing such encounters as independent environments preserves their interactive challenges while making task conditions controllable and outcomes measurable.

We introduce \gamedojo{}, a challenging benchmark of 120 boss encounters across 20 popular and classic games, together with \gamedojo{}-Data and \gamedojo{}-4B.
Gameplay controllers collect a 600-hour trajectory dataset, \gamedojo{}-Data, where we train \gamedojo{}-4B.
Built on Gymnasium, \gamedojo{} supports custom experiments and provides a unified evaluation interface for studying model capabilities, learning algorithms, and agent harnesses on interactive gameplay tasks.

Our contributions are threefold:
\begin{itemize}
  \item \textbf{A diverse, controllable and challenging benchmark for interactive intelligence.} We introduce \gamedojo{}, with 120 boss encounters across 20 diverse game environments. Configurable mechanics, action interfaces, timing modes and difficulties support controlled studies of model capabilities, learning algorithms, and agent harnesses.

  \item \textbf{An efficient and automated pipeline to collect a high-quality gameplay dataset.} We build a pipeline in which coding agents read game source code and write the behavior-tree controllers to efficiently collect gameplay data in parallel. This produces \gamedojo{}-Data, with 600 hours of high-quality gameplay trajectories covering all 120 encounters at three difficulty levels, training generalist agents and specialist policy network.

  \item \textbf{Extensive experiments on interactive performance, learning, and transfer.}
   GPT-6.1-Sol wins only 4\% of encounters at medium difficulty with time paused and 0\% in real time. Our \gamedojo{}-4B approaches frontier performance and achieves the best in real-time. Joint training improves individual tasks, and learned skills generalize across bosses and games. These findings establish \gamedojo{} as a challenging and valuable interactive gameplay benchmark for the community.
  
\end{itemize}

%% file: sections/02_related_work.tex
\section{Related Work}
\label{sec:related-work}

\noindent\textbf{Interactive decision-making and control.}
Interactive intelligence requires agents to connect perception, decisions, and actions through feedback.
Policy learning spans optimization and sequence modeling~\citep{schulman2017ppo,chen2021decisiontransformer},
while Gato and NitroGen share a policy across tasks and embodiments~\citep{reed2022gato, magne2026nitrogen}.
ReAct and Reflexion organize reasoning and memory around feedback~\citep{yao2023react,shinn2023reflexion},
showing how harness design shapes interaction.
SayCan and RT-2 connect language with robotic skills and actions~\citep{ahn2022saycan,brohan2023rt2}.
Game-TARS, SIMA-2, GameScaling and Gamer extend generalist control to virtual worlds~\citep{wang2025gametars,simateam2025sima2, zhang2026gamescaling, luo2026gamer}.
Our testbed evaluates specialist policy networks and agents built on general-purpose language models,
alongside their algorithms and harnesses.

\noindent\textbf{Game benchmarks and combat.}
Games provide repeatable challenges for evaluating interactive capabilities~\citep{zhang2026towards}.
VideoGameBench, Orak, GameVerse, and GameWorld cover diverse games and objectives~\citep{zhang2025videogamebench,park2025orak,zhang2026gameverse,ouyang2026gameworld},
while Gym-$\mu$RTS and ViZDoom provide configurable strategy and shooter environments~\citep{huang2021gymmicrorts,kempka2016vizdoom},
EvoMan studies generalist strategies across boss encounters in one game~\citep{defranca2019evoman},
and SoulsGym and DSLE focus on boss encounters in commercial games~\citep{soulsGym,gezgin2026dsle}.
Video-Arena examines state tracking from gameplay videos~\citep{zhang2026videoarena}.
Other focused work covers arcade fighting~\citep{palmas2022diambra,li2024fightladder}
and visual combat control in 3D action RPGs~\citep{chen2025combatvla}.
\gamedojo{} provides independently reconstructed boss encounters with configurable mechanics and difficulty
across 20 games in nine interaction patterns.

\noindent\textbf{Training in virtual environments.}
Virtual environments make repeated interaction practical for studying how agents learn.
ALE provides a shared suite of Atari games~\citep{bellemare2013ale},
while AlphaGo combines learning with search in Go and AlphaStar uses multi-agent learning for competitive play in StarCraft2~\citep{silver2016alphago,vinyals2019alphastar}.
VPT learns a Minecraft policy from video with inferred action labels~\citep{baker2022vpt}.
OpenAI Gym standardizes environment interfaces~\citep{brockman2016gym},
Procgen tests policy generalization across generated levels~\citep{cobbe2020procgen},
and Isaac Gym enables parallel simulation for robot learning~\citep{makoviychuk2021isaacgym}.
GameScaling studies data, diversity, and model scaling in gameplay pretraining~\citep{zhang2026gamescaling},
while Player-1 combines imitation and reinforcement learning in synthetic environments~\citep{ouyang2026player1}.
GameHorizon Suite provides gameplay data and evaluation at multiple timescales~\citep{wang2026gamehorizon}.
Our environments expose states, rewards, and mechanics for parallel training and
standardized evaluation at fixed difficulty settings.

%% file: sections/03_environments.tex
\section{\gamedojo{} Environments}
\label{sec:environments}

\subsection{Task Definition and Notation}
\label{sec:task-definition}

Each task starts at a boss encounter, with the goal of defeating the boss. At decision step $t$, the environment renders its internal state $s_t$ into a visual observation $o_t$.
The agent selects an action $a_t\sim\pi(\cdot\mid o_t, h_t)$ from its observation and optional history $h_t=(o_0,a_0,\ldots,a_{t-1})$.
The action interface executes $a_t$, and the environment advances to $s_{t+1}$ under the selected timing mode and renders the next observation.
This loop requires the agent to recognize attack cues, anticipate the boss's behavior, and adjust its strategy and control in response to feedback of the boss and environment.

\subsection{\gamedojo{} Construction}
\label{sec:game-construction}

\gamedojo{} contains 120 boss encounters across 20 environments: 10 environments and 60 encounters each in 2D and 3D.
We select representative games and bosses as references, covering the nine main interaction patterns in Table~\ref{tab:task-diversity}, from action platforming and bullet-hell dodging to first-person shooting and racing.
Differences in viewpoints, movement systems, and combat mechanics call for different strategies and forms of control. Table~\ref{tab:benchmark-comparison} compares benchmarks by environment count, task count, source availability, and core interactions.
General game benchmarks combine navigation, puzzle solving, combat, and memory in broader tasks. The specialist adversarial environments in the table each focus on one interaction family in a single game.
\gamedojo{} uses individual boss encounters as a common task unit across nine interaction patterns.

For each selected reference game and boss in Table~\ref{tab:task-diversity}, we prepare the gameplay videos, wikis, and mechanics descriptions.
The videos show attack cues, action timing, and phase changes; the wikis and mechanics descriptions detail hit rules, resource constraints, and other mechanics.
The agent first uses gameplay videos, wikis, and mechanics descriptions to identify the core elements to retain, define simplifying assumptions, and specify implementation parameters. It then implements the encounters as lightweight, standalone and high-quality environments by  Claude-Opus-5.5~\citep{anthropic2026opus55} with Claude Code at \texttt{max} reasoning effort.

\subsection{Environment Details}
\label{sec:environment-details}

Each \gamedojo{} encounter is wrapped as a Gym environment~\citep{brockman2016gym} and uses a fixed rate of 60 Hz for both rendering and simulation ticks.
The wrapper exposes \texttt{reset()} to start an episode and return the initial observation, and \texttt{step(action)} to apply an action and return the next observation, reward, and episode termination status.
\gamedojo{} has the environment source code and can expose internal states, observations, controls, and mechanics parameters as needed, or customize resets, simulation timing, and rewards. In principle, \gamedojo{} enables developers to conduct any custom experiment.
The standard \gamedojo{} evaluation uses two action interfaces, two timing modes, three difficulty settings, and a common reward definition.

\noindent\textbf{Action interfaces.}
\textbf{CUA} accepts keyboard and mouse button presses and releases, as well as mouse movement. The model outputs a structured action chunk $a_t$ that the interface parses into low-level device commands~\citep{lumine2025, omnigamearena2026}. Each chunk covers 200\,ms and uses the following format.
\begin{equation}
  a_t = \texttt{<|action\_start|>}\ \Delta x_t\,\Delta y_t\,\Delta z_t
  \texttt{;}\, K_t^{(1)}\texttt{;}\ldots\texttt{;}\, K_t^{(6)}\
  \texttt{<|action\_end|>},
\end{equation}
where $(\Delta x_t,\Delta y_t)\in[-1000,1000]^2$ are discretized relative mouse moves, $\Delta z_t\in[-5,5]$ is the scroll delta, and each $K_t^{(j)}$ contains 0-4 keys or mouse buttons held for 33\,ms.

\textbf{Semantic} maps parameterized primitives such as movement, attacks, dodges, and skills to low-level controls. For example, the model can issue a single 200\,ms action that combines moving left, jumping, and shooting.
\begin{equation}
  a_t = \texttt{\{"move": "left", "vert": "jump", "attack": "shoot 135"\}}.
\end{equation}
Both interfaces share the same visual observations and task objectives so that we can compare performance across levels of action abstraction.

\noindent\textbf{Timing modes.}
In \textbf{real-time mode}, the simulation continues during inference. In \textbf{time-paused mode}, it freezes during each decision and advances after receiving the action.
Both interfaces use a standard action interval of $0.2$ seconds. Each action advances $0.2$ seconds of simulation time in time-paused mode; in real-time mode, inference latency affects how often the agent acts.
Two modes compare decision-making and execution under different time constraints.

\noindent\textbf{Difficulty and reward.}
Each encounter has fixed Easy, Medium, and Hard parameter sets, calibrated as described in Section~\ref{sec:scripted-collection}.
All encounters use the same reward definition: given the boss's remaining health $H_t$ in state $s_t$ and its total health $H_{\max}$ in the current configuration, the step reward is $r_t=(H_t-H_{t+1})/H_{\max}$.
Evaluation reports average win rate and average progress: whether the boss is defeated and the fraction of its health removed, respectively (Section~\ref{sec:settings}).

\subsection{Script Design and Dataset Collection}
\label{sec:scripted-collection}

We use behavior-tree controllers that read internal environment state as reference policies for calibrating difficulty.
We give Claude-Opus-5.5~\citep{anthropic2026opus55} with Claude Code the environment source code, state information, and task instructions, and have it write a behavior tree to control the player character in each encounter.
We add a fixed $200$ ms delay to the controller's responses to approximate human reaction time.

We test all three configurations for each encounter with the same script and tune the parameters against target defeat times: about $20$ seconds on Easy, about $60$ seconds on Medium, and several minutes on Hard.
At higher difficulties, bosses deal more damage, have more health and skills, and are more likely to initiate attacks with less reaction time.

Prior gameplay datasets have relied heavily on human recordings, which can include failed attempts and actions unrelated to the target task.
These lower-quality demonstrations can hurt imitation learning. On Q*bert, behavior cloning on the top-scoring 5\% of human episodes outperformed training on the full dataset~\citep{kanervisto2020benchmarking}.
Cleaning such recordings at scale takes substantial manual effort. Yue et al.~\citep{yue2026scalingbc} manually review flagged clips and spot-check additional recordings, a process that can still miss low-quality segments.

We collect \gamedojo{}-Data by running the behavior-tree controllers in parallel. Once the controllers are in place, adding simulator instances increases collection throughput without recruiting more players.
Each controller targets a specific boss, so we can record demonstrations aligned with the encounter objective and verify episode outcomes from the game state.
We collected 600 hours of trajectories for fine-tuning across all 120 encounters and three difficulty levels, averaging 100 minutes per encounter and difficulty.

%% file: sections/04_evaluation.tex
\section{Main Experiments}
\label{sec:evaluation}
\label{sec:experiments}

We evaluate \gamedojo{} through both existing capabilities and learning from interaction.
We first compare frontier models across comprehensive evaluations, then examine how \gamedojo{}-Data improves agent performance and how specialist policies learn from scratch.

\gamedojoDifficultyTable

\subsection{Experimental Settings}
\label{sec:settings}

\noindent\textbf{Models.} We evaluate eleven frontier models: GPT-6.1-Sol~\citep{openai2026sol}, GPT-6-Luna~\citep{openai2026luna},
Claude-Opus-5.5~\citep{anthropic2026opus55}, Claude-Haiku-5.5~\citep{anthropic2026haiku55}, Claude-Haiku-4.5~\citep{anthropic2025haiku45},
Gemini-3.8-Flash~\citep{google2026gemini38flash}, DeepSeek-V4.1-Flash~\citep{deepseek2026v41},
GLM-5.3-Flash~\citep{zai2026glm53flash}, MiniMax-M3~\citep{minimax2026m3}, and Qwen3.5-35B-A3B and Qwen3.5-4B~\citep{qwen2026qwen35}.
We also include GameScaling-9B-S3~\citep{zhang2026gamescaling} and Gamer-35B-A3B~\citep{luo2026gamer} trained on diverse gameplay trajectories as two generalist game model comparison.
We include three baselines: Idle issues no actions, while Random samples actions from the available action space at each decision step.
Boss-specific script uses the state-aware controllers written by coding agents in Section~\ref{sec:scripted-collection}, providing a privileged upper-bound reference.
\gamedojo{}-4B denotes Qwen3.5-4B after supervised fine-tuning on \gamedojo{}-Data. Transfer experiments independently fine-tune Qwen3.5-4B on trajectories from specified source bosses.
We also evaluate PPO visual policies trained from scratch~\citep{schulman2017ppo}, following  DSLE~\citep{gezgin2026dsle}.

\noindent\textbf{Tasks and interaction.}
Agents must avoid death and deal as much damage as possible to the boss within a fixed interaction budget.
Time-paused episodes allow 300 steps and can take tens of minutes end to end. Real-time episodes allow 180 s including inference.
Generalist models receive the two most recent frames and actions, while specialist policy networks observe only the current frame.

\noindent\textbf{Reasoning budget.}
Language-model agents use \texttt{low} reasoning where supported.
Each model retains the same reasoning setting across all four interaction conditions.
The supplementary material records the settings and output limits.

\noindent\textbf{Metrics.}
We report average win rate (WR) and average progress (PG).
For case $i$, let $w_i=1$ if the boss is defeated and $0$ otherwise; progress is the fraction of total boss health removed.
For a fixed model, interface, timing mode, and difficulty, we average over all $N$ cases:
\begin{equation}
  \mathrm{WR}=\frac{1}{N}\sum_{i=1}^{N}w_i,
  \qquad
  \mathrm{PG}=\frac{1}{N}\sum_{i=1}^{N}\frac{\Delta H_i}{H_i^{\mathrm{total}}}.
\end{equation}
Here, $\Delta H_i$ is the boss health removed and $H_i^{\mathrm{total}}$ is its total health.
WR measures how often agents defeat the boss; PG captures partial progress within an encounter.
Even when WR is zero, PG distinguishes models by how far they progress through an encounter.
All cases receive equal weight and difficulty levels are reported separately.
Tables report WR / PG in percentages; training curves show episode progress in $[0,1]$. More details are in supplement material.

\subsection{Frontier Models across Interaction Conditions}
\label{sec:main-results}
\label{sec:main-frontier}
We begin by evaluating frontier models on all 120 bosses in \gamedojo{} at medium difficulty.
Each of the eleven models runs under time-paused Semantic, time-paused CUA, real-time Semantic, and real-time CUA conditions, yielding $48$ model configurations.
Table~\ref{tab:frontier-main} groups results by timing mode and interface, with WR / PG over all 120 bosses.

\noindent\textbf{Strong models make progress but rarely win.}
In paused mode, GPT-6.1-Sol and Claude-Opus-5.5 achieve the highest average PG at 44.77\% and 31.23\%, respectively, compared with 7.66\% for Random, but they only win 4.6\% and even 0\%, showing the difficulty to completely beat a boss. GPT-6.1-Sol survives until the end in 47.6\% and 1.7\% of encounters in time-paused and real-time settings, showing that it can avoid boss attacks when having time to reason.
It still cannot match the scripts' ability to dodge precisely while maximizing damage output, which explains the large gap in scores.
The results show that \gamedojo{} remains far from saturation even in paused mode.

\noindent\textbf{Weaker models perform close to or below Random.}
Claude-Haiku-4.5 achieves 9.28\% average PG in paused mode, only slightly above Random's 7.66\%, with zero WR.
Under paused CUA, Qwen3.5-4B reaches just 1.94\% PG, below Random's 6.81\%.
Even GameScaling-9B-S3, a generalist model pretrained on gameplay trajectories~\citep{zhang2026gamescaling}, reaches only 5.28\% PG and zero WR in the same setting.
Boss encounters remain challenging for smaller generalist models, even for those already trained on diverse gameplay.

\noindent\textbf{Most models make similar progress with Semantic and CUA.}
In paused mode, the median absolute PG gap between the two interfaces is 1.6 percentage points across the eleven frontier models.
Gemini-3.8-Flash reaches 31.1\% PG with Semantic and 21.0\% with CUA, while GLM-5.3-Flash reaches 23.4\% with CUA and 18.9\% with Semantic.

\noindent\textbf{All models fall below Random in real time.}
In real time, all evaluated models score below Random in PG and achieve zero WR.
GPT-6.1-Sol's average PG drops from 44.77\% to 2.70\%, while Opus-5.5 drops from 31.23\% to 1.19\%.
In a medium-difficulty CUA diagnostic with one episode per boss, switching from time-paused to real-time interaction reduces the mean number of recorded decisions per episode from 222.3 to 6.7 for GPT-6.1-Sol and from 229.1 to 5.4 for Claude-Opus-5.5.
Both fall well below Random.
GameScaling-9B-S3 also reaches only 1.50\% PG.
The sharp decline across both interfaces shows that acting effectively while the environment continues to change remains a major challenge.
The state-aware boss-specific script reaches 99.83\% WR and 99.99\% PG, showing the large gap to strong task-specific control.

\subsection{Learning from Gameplay Trajectories}
\label{sec:main-sft}

After assessing frontier capabilities, we examine how trajectory supervision improves interactive performance.
We obtain \gamedojo{}-4B by supervised fine-tuning Qwen3.5-4B on 240 hours sampled uniformly from \gamedojo{}-Data: 40 minutes for each boss at each of three difficulty levels.

Table~\ref{tab:difficulty} shows large gains from trajectory supervision across all three difficulty levels.
In paused mode, \gamedojo{}-4B raises average PG from 3.04\% to 42.92\% and average WR from 0.09\% to 21.39\%.
\gamedojo{}-4B reaches performance comparable to GPT-6.1-Sol: leading 0.83\% in WR while losing 2.36\% in PG, respectively.
In real time, \gamedojo{}-4B achieves the best performance among all models at every difficulty level, reaching 26.26\%, 8.18\%, and 2.48\% on easy, medium, and hard encounters.
Across the three difficulty levels and both timing modes, the share of recorded decisions that pass CUA action format validation rises from 8.94\%--11.88\% before SFT to 99.96\%--99.98\% after SFT.
These gains show that scripted gameplay trajectories therefore provide a promising source of training data, and offers an effective way to improve interactive agents.

\subsection{Learning with Specialist Policies}
\label{sec:main-ppo}
Specialist policy networks offer another view of task difficulty by learning a mapping from visual observations to actions directly from environment feedback.
We use Terraria and Doom as representative 2D and 3D environments.
For each environment, we train six separate single-boss PPO policies and one shared policy on a mixture of all six bosses.
Both training and evaluation use easy difficulty, time-paused interaction. Outputs use each game's Semantic interface; rewards and action intervals follow \gamedojo{}.

\noindent\textbf{Boss encounters pose distinct learning challenges.}
Figure~\ref{fig:ppo-boss-learning} shows different learning curves across bosses even with the same initialization.
In Terraria, PG approaches 100\% on Eye of Cthulhu and Eater of Worlds, while King Slime and Queen Bee approache 50\%.
In Doom, PG on Cyberdemon stays near 0\%, and Barons, Archvile, and Mancubus remain below approximately 10\% late in training.
These differences reveal a range of encounter difficulties under a shared embodiment: even the easy configuration remains challenging for specialist policies trained with PPO.

%% file: sections/05_experiments.tex
\section{Further Analysis}
\label{sec:further-analysis}

\gamedojoBossCheckpointFigures

We extend the evaluation by examining the difficulty separation, the generalization across the embodiment, boss and game, and the comparison of separate and joint training.

\subsection{Performance across Difficulty Levels}
\label{sec:analysis-difficulty}
We examine the difficulty gradient across easy, medium, and hard configurations on the same tasks.
Table~\ref{tab:difficulty} compares different models on all 120 bosses through CUA.

\noindent\textbf{Performance drops consistently with difficulty.}
PG decreases from easy to medium to hard for every evaluated model in both timing modes.
For example, \gamedojo{}-4B's paused PG falls from 76.70\% to 40.32\% and 11.75\% and GPT-6.1-Sol falls from 78.08\% to 44.59\% and 13.17\%.
All reported model WRs on hard are zero in both modes, showing that the difficulty gradient extends to encounters that remain unsolved even after trajectory supervision.

\noindent\textbf{Real-time interaction remains demanding even on easy.}
On easy encounters, switching to real time reduces GPT-6.1-Sol's PG from 78.08\% to 13.65\% and \gamedojo{}-4B's from 76.70\% to 26.26\%.
On hard encounters, the highest real-time PG among learned agents is only 2.48\%, close to Random's 2.31\%.
The boss-specific script reaches 35.83\% WR and 93.23\% PG under the same 180-second budget, exposing a large gap to the privileged reference.
Effective control under continuous environment changes remains challenging across difficulty levels.

\subsection{Learning and Transfer across Bosses}
\label{sec:analysis-boss-transfer}

We compare six-boss joint training and fixed-source transfer in Terraria and Doom using PPO and SFT.
The joint and single-boss runs use the same training budget per boss within each method (100K online steps for PPO and 2 hours of demonstrations for SFT).

\noindent\textbf{Joint training improves mean progress on individual bosses.}
At the final checkpoint in Figure~\ref{fig:boss-joint-checkpoints}, six-boss joint PPO raises mean PG over separate training by 7.51 percentage points in Terraria and 26.46 in Doom.
Across all twelve bosses, joint PPO raises mean PG from 47.7\% to 64.7\%. Joint SFT raises it from 69.5\% to 87.5\%.

\noindent\textbf{Single-boss training transfers within each game.}
We train on Queen Bee in Terraria or Cyberdemon in Doom and evaluate on the other five bosses in the same game (Figure~\ref{fig:boss-transfer-checkpoints}).
From initialization to the final checkpoint, fixed-source PPO raises mean PG by 34.96 percentage points in Terraria and 4.81 in Doom, going from 28.3\% to 48.2\% across the ten held-out bosses.
The larger joint-training gain occurs in Doom, and the larger fixed-source gain occurs in Terraria.
Across the same ten bosses, fixed-source SFT raises mean PG from 9.5\% to 68.7\%, finishing 6.8 percentage points below direct training on each target boss.

\subsection{Transfer across Games}
\label{sec:analysis-game-transfer}

Fixed-source SFT reaches 73.65\% mean PG on the other five Terraria bosses and 63.81\% on the other five Doom bosses (Figure~\ref{fig:boss-transfer-checkpoints}).
We fine-tune Qwen3.5-4B on \gamedojo{}-Data from one or six bosses in Terraria or Doom, or twelve bosses together. \gamedojo{}-4B provides the all-boss comparison. Figure~\ref{fig:cross-game-transfer} reports PG gains over Qwen3.5-4B for six model settings on five boss sets.

\noindent\textbf{Joint training yields the largest mean PG gains.}
Combining both games extends these gains beyond the source environments: joint SFT outperforms the four single-game variants in all five evaluation sets.
PG improves by 15.39 and 14.48 percentage points on unseen 2D and 3D games, respectively, averaging 14.94 points over their 108 bosses.

\noindent\textbf{Transfer depends on the source environment.}
Both Doom-only models gain across all five groups, including unseen 2D games.
Terraria-only models gain 4.19--5.77 points on other 2D games but lose 1.05--5.10 points on other 3D games.
Adding Terraria bosses improves within-game performance without removing this directional difference, highlighting the value of diverse source environments for broader transfer.

%% file: sections/06_discussion.tex
\section{Conclusion and Limitations}
\label{sec:conclusion}
\label{sec:limitations}
\textbf{Conclusion.} We present \gamedojo{}, a controllable and challenging interactive benchmark of 120 boss encounters across 20 games, together with the 600-hour \gamedojo{}-Data, the automatic data collection pipeline and the trajectory-supervised \gamedojo{}-4B.
These resources support controlled studies of model capabilities, learning algorithms, and agent harnesses for interactive intelligence.

\textbf{Limitations.}
Reconstructing encounters from videos, wikis and mechanics descriptions may omit implicit rules and visual details of the original games.
Evaluation therefore measures performance in the reconstructed tasks; transfer to the original games and physical environments requires extra evaluation in those settings.

\noindent\textbf{Copyright and use.} All game titles, characters, artwork and other intellectual property referenced on this page belong to their respective owners and publishers.
GameDojo is not affiliated with, endorsed or sponsored by any of them.
The encounters are independent reconstructions made for non-commercial scientific research only; game covers and screenshots are shown solely to identify the reference titles.
During reconstruction, the models used public gameplay videos, wikis and descriptions of game mechanics as references and had no access to the original games' source code.
We did not reverse engineer or crack the original games.
GameDojo research materials are for non-commercial use.

%% file: sections/07_appendix.tex
\providecommand{\gamedojoMainRef}[2]{#2}
\providecommand{\gamedojoSupplementFloatBarrier}{}

\section{Environment Construction and Design}
\label{sec:app-env}


\subsection{Games and Boss Encounters}
\label{sec:app-catalog}

\begingroup
\setlength{\parskip}{3pt}
\interlinepenalty=10000
\gamedojo{} spans nine interaction patterns across ten 2D and ten 3D environments, with six encounters per game.

\begin{table*}[!htbp]
\centering
\scriptsize
\setlength{\tabcolsep}{5pt}
\renewcommand{\arraystretch}{1.12}
\caption{\textbf{Full \gamedojo{} Environment Catalog.} All 20 games (10 2D and 10 3D) and their 6 boss encounters (120 encounters in total).}
\label{tab:game-catalog}
\begin{adjustbox}{max width=\linewidth}
\begin{tabular}{@{}ll>{\raggedright\arraybackslash}p{8.85cm}@{}}
\toprule
\textbf{Game} & \textbf{Viewpoint \& Genre} & \textbf{Six Boss Encounters} \\
\midrule
\multicolumn{3}{@{}l}{\textit{2D Environments (10 Games, 60 Encounters)}} \\
\emph{Hollow Knight} & 2D Action Platformer & Hollow Vessel, Iron Bulwark, Lumen Moth, Needle Warden, Pale Vessel, Sickle Sisters \\
\emph{Kirby} & 2D Side-Scrolling Platformer & Kaboola, King Dedede, Kracko, Lololo \& Lalala, Poppy Bros.\ Sr., Whispy Woods \\
\emph{Terraria} & 2D Sandbox Action & Brain of Cthulhu, Eater of Worlds, Eye of Cthulhu, King Slime, Queen Bee, Skeletron \\
\emph{The King of Fighters '97} & 2D Fighting & Iori, Orochi, Orochi Chris, Orochi Iori, Orochi Shermie, Orochi Yashiro \\
\emph{Battle City} & 2D Top-Down Tank Defense & Brick Front, Jungle Ambush, River Delta, River Ice, Steel Fort, Steel Maze \\
\emph{Cuphead} & 2D Run-and-Gun Boss Rush & Cagney Carnation, Goopy Le Grande, Hilda Berg, Ribby \& Croaks, The Root Pack, Wally Warbles \\
\emph{Touhou} & 2D Vertical Bullet Hell & Cirno, Hong Meiling, Patchouli Knowledge, Remilia Scarlet, Rumia, Sakuya Izayoi \\
\emph{Soul Knight} & 2D Top-Down Dungeon Crawler & Christmas Treant, Devil's Snare, Goblin Priest, Grand Slime, Varkolyn Leader, Volcanic Sandworm \\
\emph{Mega Man} & 2D Precision Platformer & Bomb Man, Cut Man, Elec Man, Fire Man, Guts Man, Ice Man \\
\emph{The Binding of Isaac} & 2D Top-Down Twin-Stick & Duke of Flies, Envy, Famine, Gemini, Larry Jr., Monstro \\
\midrule
\multicolumn{3}{@{}l}{\textit{3D Environments (10 Games, 60 Encounters)}} \\
\emph{Minecraft} & 3D First-Person Sandbox & Ender Dragon, Iron Golem, Skeleton Archer, Warden, Wither, Zombie Leader \\
\emph{Doom} & 3D First-Person Shooter & Arch-vile, Barons of Hell, Cyberdemon, Mancubus, Revenant, Spider Mastermind \\
\emph{Sekiro} & 3D Third-Person Sword Action & Demon of Hatred, Genichiro, Guardian Ape, Gyoubu, Isshin, Lady Butterfly \\
\emph{Genshin Impact} & 3D Third-Person Party Action & Andrius, Electro Hypostasis, Maguu Kenki, Oceanid, Ruin Guard, Signora \\
\emph{Quake II} & 3D First-Person Shooter & Gladiator, Hornet, Makron, Mutant, Supertank, Tank Commander \\
\emph{Hexen II} & 3D First-Person Fantasy & Airborne Bone Rider, Axe-Hurling Rider, Blinking Tether Rider, Boar Archer Rider, Orb-Warded Serpent, Star-Volley Sorcerer \\
\emph{Super Mario 64} & 3D Platformer & Big Bully, Bowser in the Dark World, Eyerok, King Bob-omb, Whomp King, Wiggler \\
\emph{Tomb Raider} & 3D Third-Person Action & Cannon Quadruped, Giant Lizard, Legless Giant, Lone Gunman, Twin Pistols, Winged Matriarch \\
\emph{Zelda: Ocarina of Time} & 3D Third-Person Lock-On Action & Armored Spider Queen, Burrowing Fire Dragon, Lava Lizard King, Specter Rider, Twin Flame-Frost Witches, Water-Core Amoeba \\
\emph{Need for Speed: Most Wanted} & 3D Arcade Racing & Drag, Knockout, Pursuit, Speedtrap, Street Champion, Tollbooth \\
\bottomrule
\end{tabular}
\end{adjustbox}
\end{table*}

\noindent\textbf{(a) Battle City}~\citep{namco1985battlecity}.
In this top-down tank game, terrain shapes the firing lanes. Players line up shots from cover and move before enemy fire cuts off their route.\\*
\emph{Encounters.} Brick Front (brick cover), Jungle Ambush (concealed approaches), River Delta (water crossings), River Ice (icy lanes), Steel Fort (steel fortifications), Steel Maze (steel-walled corridors).

\noindent\textbf{(b) Cuphead}~\citep{studiomdhr2017cuphead}.
This run-and-gun game pairs sustained fire with jumps and dashes through patterned attacks. In aerial encounters the player can also change altitude while firing.\\*
\emph{Bosses.} Cagney Carnation (projectile-spitting flower), Goopy Le Grande (bouncing slime), Hilda Berg (aerial transformations), Ribby \& Croaks (frog duo), The Root Pack (vegetable trio), Wally Warbles (airborne bird).

\noindent\textbf{(c) Hollow Knight}~\citep{teamcherry2017hollowknight}.
This 2D action platformer centers on close-range sword combat. Players read windups and use jumps or dashes to get into striking range as an attack ends.\\*
\emph{Bosses.} Hollow Vessel (vessel duelist), Iron Bulwark (armored heavyweight), Lumen Moth (luminous flier), Needle Warden (needle-wielding duelist), Pale Vessel (pale swordfighter), Sickle Sisters (paired sickles).

\noindent\textbf{(d) Kirby's Dream Land}~\citep{hallaboratory1992kirby}.
This side-scrolling platformer lets players hover to change their height and avoid attacks. Grounded and airborne bosses call for different approaches to getting in range.\\*
\emph{Bosses.} Kaboola (armed blimp), King Dedede (hammer wielder), Kracko (lightning cloud), Lololo \& Lalala (block-pushing duo), Poppy Bros.\ Sr.\ (bomb thrower), Whispy Woods (apple-dropping tree).

\noindent\textbf{(e) The King of Fighters '97}~\citep{snk1997kof}.
In these one-on-one bouts, players manage spacing, block or evade attacks, and counter during recovery.\\*
\emph{Bosses.} Iori (purple flames), Orochi (screen-wide blasts), Orochi Chris (flame pressure), Orochi Iori (fast swipes), Orochi Shermie (lightning), Orochi Yashiro (grappling).

\noindent\textbf{(f) Mega Man}~\citep{capcom1987megaman}.
Players in this 2D action platformer jump over attacks while lining up horizontal shots. Each Robot Master changes when and where the player can safely fire.\\*
\emph{Bosses.} Bomb Man (arcing bombs), Cut Man (returning cutters), Elec Man (electric beams), Fire Man (flame waves), Guts Man (thrown boulders), Ice Man (ice projectiles).

\noindent\textbf{(g) Soul Knight}~\citep{chillyroom2017soulknight}.
This top-down shooter requires players to aim while moving out of projectile lanes. Boss summons and moving hazards can change which part of the arena is safe.\\*
\emph{Bosses.} Christmas Treant (charging tree), Devil's Snare (projectile-spitting plant), Goblin Priest (spell volleys), Grand Slime (jumping slime), Varkolyn Leader (poison attacks), Volcanic Sandworm (burrowing ambushes).

\noindent\textbf{(h) The Binding of Isaac}~\citep{mcmillen2011isaac}.
In this top-down shooter, movement and firing direction are independent. Players can retreat while shooting, but splitting enemies and sweeping attacks can take away the space they need to dodge.\\*
\emph{Bosses.} Duke of Flies (fly swarm), Envy (splitting enemy), Famine (mounted charges), Gemini (paired attackers), Larry Jr.\ (segmented pursuit), Monstro (leaps and projectile bursts).

\noindent\textbf{(i) Terraria}~\citep{relogic2011terraria}.
These 2D sandbox combat encounters require players to use vertical space and fire while moving. The boss's trajectory determines whether the player should keep distance, change elevation, or cross the arena.\\*
\emph{Bosses.} Brain of Cthulhu (teleporting brain), Eater of Worlds (segmented burrower), Eye of Cthulhu (charging eye), King Slime (leaping and teleporting slime), Queen Bee (dashes and stingers), Skeletron (head and independently attacking hands).

\noindent\textbf{(j) Touhou: Embodiment of Scarlet Devil}~\citep{teamshanghaialice2002touhou}.
In this vertical bullet-hell shooter, players thread gaps in dense barrages while keeping their shots on the boss.\\*
\emph{Bosses.} Cirno (ice barrages), Hong Meiling (colorful spreads), Patchouli Knowledge (elemental barrages), Remilia Scarlet (scarlet patterns), Rumia (darkness attacks), Sakuya Izayoi (knife volleys).

\noindent\textbf{(k) Doom}~\citep{idsoftware1993doom}.
This first-person shooter combines continuous aiming with strafing around projectiles and moving enemies. The encounters vary how much time and space players have to stay on target.\\*
\emph{Bosses.} Arch-vile (flame attack), Barons of Hell (paired plasma throwers), Cyberdemon (rocket volleys), Mancubus (spread fireballs), Revenant (homing missiles), Spider Mastermind (rapid gunfire).

\noindent\textbf{(l) Genshin Impact}~\citep{mihoyo2020genshin}.
These third-person action RPG encounters alternate between avoiding attack patterns and closing in to deal damage. Changing phases, summons, and protected targets alter when the player can attack effectively.\\*
\emph{Bosses.} Andrius (wolf lunges), Electro Hypostasis (exposed core), Maguu Kenki (phantom sword strikes), Oceanid (water summons), Ruin Guard (spinning automaton), Signora (ice and fire phases).

\noindent\textbf{(m) Hexen II}~\citep{ravensoftware1997hexen}.
In this first-person fantasy action game, players aim at mobile targets while dodging spells and thrown weapons. Flying, mounted, and teleporting enemies change where the player needs to look next.\\*
\emph{Bosses.} Airborne Bone Rider (aerial rider), Axe-Hurling Rider (thrown axes), Blinking Tether Rider (teleporting tether attacks), Boar Archer Rider (mounted archery), Orb-Warded Serpent (protective orbs), Star-Volley Sorcerer (star volleys).

\noindent\textbf{(n) Minecraft}~\citep{mojang2011minecraft}.
These first-person sandbox encounters require players to track mobs in three dimensions and control their distance. Some enemies can be engaged at range, while others force the player to reposition before getting a clear attack.\\*
\emph{Bosses.} Ender Dragon (aerial passes), Iron Golem (heavy melee swings), Skeleton Archer (ranged arrows), Warden (sonic attack), Wither (explosive skulls), Zombie Leader (undead melee pressure).

\noindent\textbf{(o) Need for Speed: Most Wanted}~\citep{eacanada2005mostwanted}.
The racing environment requires players to coordinate steering, braking, and acceleration. Its events change what counts as progress, from holding speed through checkpoints to finishing ahead of a rival or escaping pursuit.\\*
\emph{Events.} Drag (short sprint), Knockout (last-place elimination), Pursuit (police evasion), Speedtrap (speed at checkpoints), Street Champion (rival race), Tollbooth (timed checkpoints).

\noindent\textbf{(p) Quake II}~\citep{idsoftware1997quake2}.
In this first-person shooter, players track moving targets while strafing or breaking line of sight. The encounters range from fast lunging enemies to large machines with sustained fire.\\*
\emph{Bosses.} Gladiator (railgun shots), Hornet (flying gunship), Makron (heavy weapon fire), Mutant (leaping melee), Supertank (armored gun platform), Tank Commander (armored ranged attacker).

\noindent\textbf{(q) Sekiro}~\citep{fromsoftware2019sekiro}.
This third-person sword combat game rewards deflecting at the moment of impact. Players read attack sequences to decide when to deflect, dodge, or counter.\\*
\emph{Bosses.} Demon of Hatred (wide fire attacks), Genichiro (sword and bow), Guardian Ape (sweeping attacks), Gyoubu (mounted spear), Isshin (multi-weapon duel), Lady Butterfly (aerial attacks).

\noindent\textbf{(r) Super Mario 64}~\citep{nintendo1996mario64}.
These 3D platforming encounters make position and momentum part of the attack. Players may need to get behind an enemy, redirect it toward a hazard, or land on a moving target.\\*
\emph{Bosses.} Big Bully (knockback near lava), Bowser in the Dark World (tail-grab throws), Eyerok (eye-bearing stone hands), King Bob-omb (grab-and-throw duel), Whomp King (ground-pound opening), Wiggler (stomp and pursuit).

\noindent\textbf{(s) The Legend of Zelda: Ocarina of Time}~\citep{nintendo1998ocarina}.
These third-person encounters revolve around finding weak points and acting while they are exposed. Players reposition through arena hazards to reach the right angle or range.\\*
\emph{Bosses.} Armored Spider Queen (armored arachnid), Burrowing Fire Dragon (emerging from below), Lava Lizard King (lava-arena lizard), Specter Rider (mounted apparition), Twin Flame-Frost Witches (fire and ice attacks), Water-Core Amoeba (waterborne core).

\noindent\textbf{(t) Tomb Raider I}~\citep{coredesign1996tombraider}.
This third-person action adventure requires players to move while firing and keep enemies in view. Grounded and flying opponents differ in how they close distance and pressure the player.\\*
\emph{Bosses.} Cannon Quadruped (four-legged gunner), Giant Lizard (reptilian attacker), Legless Giant (torso mutant), Lone Gunman (solo shooter), Twin Pistols (dual-pistol shooter), Winged Matriarch (aerial attacker).

\par\endgroup

Table~\ref{tab:game-catalog} lists each game, its viewpoint and genre, and all six boss encounters.

\gamedojoSupplementFloatBarrier
\subsection{Coding Agent Game Reconstruction}
\label{sec:app-reconstruction}

\noindent\textbf{Macro reconstruction workflow.}
We construct all 20 games from scratch using Claude-Opus-5.5 through Claude Code with \texttt{max} reasoning effort.
For each target game, we gather public gameplay videos, wiki pages, and mechanics summaries, then run a three stage workflow:
\begin{enumerate}
  \item \emph{Specification writing}: The coding agent writes a structured mechanics document covering player movement, attacks, cooldowns, resource rules, boss attack phases, hitboxes, visual cues, and win or loss conditions for all six bosses.
  \item \emph{Engine and boss implementation}: The agent implements the game simulation, arena geometry, boss state machines, and visual renderer using lightweight assets that keep combat cues readable without copying commercial files.
  \item \emph{Three round verification loop}: Every new game passes three test and fix rounds. Round~1 checks complete combat mechanics and boss phase transitions; Round~2 reviews visual clarity, attack warnings, and controls against reference material; Round~3 tests multi seed stability, exact replay, public Gym compliance, and frame rate. Any remaining bugs are fixed and retested before a game enters the benchmark.
\end{enumerate}

\noindent\textbf{Gym packaging and rendering backends.}
All 20 games are exposed through a single Gymnasium wrapper (\texttt{reset(seed, options)} and \texttt{step(action)}) that hides three lightweight runtime backends:
(1)~a shared 2D Canvas engine (\texttt{BB.Case}) for 7 classic 2D games (\emph{Hollow Knight}, \emph{Kirby}, \emph{Terraria}, \emph{KOF '97}, \emph{Battle City}, \emph{Cuphead}, and \emph{Mega Man}),
(2)~standalone browser based JavaScript engines for 12 games using HTML5 Canvas 2D (\emph{Touhou}, \emph{Soul Knight}, \emph{Isaac}), CPU raycasting (\emph{Doom}), or offline Three.js WebGL 3D (\emph{Sekiro}, \emph{Genshin Impact}, \emph{Quake II}, \emph{Hexen II}, \emph{Super Mario 64}, \emph{Tomb Raider}, \emph{Zelda: OoT}, and \emph{NFS: Most Wanted}), and
(3)~a Python and NumPy simulation engine with headless hardware accelerated ModernGL perspective rendering for \emph{Minecraft}.
Visual effects and camera rendering read the simulation state without modifying physics or random number streams.
For all 3D games, the wrapper enforces true perspective rendering and raises an error if hardware acceleration is unavailable, preventing software top down fallbacks from entering evaluation or training.
The public interface returns only RGB frames and public control descriptions; internal health values, hitboxes, boss move tables, and reward verifiers stay hidden inside the environment wrapper.

\noindent\textbf{Fixed game parameters.}
Every environment shares the following fixed runtime parameters:
\begin{itemize}
  \item \textbf{Resolution}: $854\times480$ pixels (16:9), \texttt{uint8} RGB array of shape $(480, 854, 3)$.
  \item \textbf{Simulation and rendering clock}: Fixed $60\text{ Hz}$ tick rate ($16.67\text{ ms}$ per tick), unified across all 20 games.
  \item \textbf{Decision step}: $200\text{ ms}$ per environment step ($5\text{ Hz}$), advancing $12$ simulation ticks divided into $6$ equal input substeps of $33.3\text{ ms}$ ($2$ ticks per substep).
  \item \textbf{Episode limit}: Hard engine timeout of $300\text{ s}$ ($18{,}000$ ticks), ending early upon player death or boss defeat on the exact terminal tick.
  \item \textbf{Determinism}: Seeded pseudo random number generators and full engine state hashing guarantee bit exact replay given the same seed, difficulty, and action sequence.
  \item \textbf{Step reward}: Normalized boss health reduction per step, $r_t = (H_t - H_{t+1}) / H_{\max}$, with cumulative progress clipped to $[0, 1]$ at episode end.
\end{itemize}

\subsection{Action Spaces}
\label{sec:app-actions}

Every environment supports two action interfaces over the same $200\text{ ms}$ decision step: a unified Computer Use Agent (\textbf{CUA}) keyboard and mouse interface and a game specific \textbf{Semantic} discrete interface.

\noindent\textbf{CUA action format.}
A CUA action is a single line token sequence specifying mouse motion and six $33.3\text{ ms}$ keyboard substeps for the next $200\text{ ms}$:
\begin{center}
\footnotesize
\texttt{<|action\_start|>}\,$X\,Y\,Z$\,\texttt{;}\,$K^{(1)}$\,\texttt{;}\,$\dots$\,\texttt{;}\,$K^{(6)}$\,\texttt{<|action\_end|>}
\end{center}
Here $X, Y \in [-1000, 1000]$ are integer relative mouse displacements ($X>0$ right, $Y>0$ down), and $Z \in [-5, 5]$ is the scroll wheel delta ($Z>0$ scroll up).
Each substep $K^{(m)}$ ($m=1,\dots,6$) lists $0$ to $4$ simultaneously held controls separated by spaces from the shared 14-control set \texttt{\{W, A, S, D, Space, Shift, J, K, L, Q, E, R, LB, RB\}}, where \texttt{LB} and \texttt{RB} denote left and right mouse buttons.
An empty substep releases all keys while keeping its semicolon separator.
A key appearing in consecutive substeps (or across adjacent $200\text{ ms}$ packets) remains continuously held rather than triggering a fresh press; a new press fires only on a released-to-held transition.
Mouse motion follows one of three per game modes: \texttt{None} (mouse motion ignored), \texttt{Pointer} (2D crosshair movement normalized to screen coordinates), or \texttt{Camera} (3D yaw and pitch camera rotation).

\noindent\textbf{Semantic action format.}
A Semantic action is a multi-head discrete choice written as a JSON dictionary mapping each action head to one named option, for example \texttt{\{"move": "left", "vert": "jump", "attack": "shoot 135"\}} in \emph{Terraria} or \texttt{\{"move": "forward", "turn": "left\_10", "attack": "fire"\}} in \emph{Doom}.
Any omitted head defaults to option~0 (\texttt{"none"}), and every combination translates deterministically into one $200\text{ ms}$ CUA packet (held across the full step).
Across 19 games, heads are defined in \texttt{compact\_actions.json} such that separate heads never share keys or mouse axes and the total number of combinations stays strictly below $1{,}000$.
\emph{Terraria} uses a three-head layout (\texttt{move} $\times$ \texttt{vert} $\times$ \texttt{attack} $= 3 \times 3 \times 19 = 171$) covering 16 directional bow angles at $22.5^\circ$ increments plus two sword swings.
Malformed strings or out of range values in either interface execute a neutral no op packet and increment the format error counter.

Table~\ref{tab:action-spaces} details both action spaces for all 20 games.

\begin{table*}[!htbp]
\centering
\scriptsize
\setlength{\tabcolsep}{3.5pt}
\renewcommand{\arraystretch}{1.14}
\caption{\textbf{Action Spaces Across All 20 Games.} For each game, we list the CUA mouse mode, active keyboard and mouse bindings with their gameplay functions, and the Semantic multi-head discrete structure with total action count.}
\label{tab:action-spaces}
\begin{adjustbox}{max width=\linewidth}
\begin{tabular}{@{}ll>{\raggedright\arraybackslash}p{7.4cm}>{\raggedright\arraybackslash}p{4.8cm}@{}}
\toprule
\textbf{Game} & \textbf{Mouse} & \textbf{CUA Controls (Key: Function)} & \textbf{Semantic Heads \& Total Size} \\
\midrule
\emph{Battle City} & None & \texttt{WASD}: 4-way tank drive; \texttt{J/K/Space/LB}: fire shell & \texttt{move(57) $\times$ fire(3) = 171} \\
\emph{Cuphead} & None & \texttt{WASD}: move/aim/crouch; \texttt{Space}: jump/parry; \texttt{Shift}: dash; \texttt{J/LB}: fire; \texttt{K/RB}: EX shot; \texttt{L}: lock aim; \texttt{Q}: swap plane gun/bomb; \texttt{E}: super & \texttt{move(28) $\times$ act(29) = 812} \\
\emph{Doom} & Camera & \texttt{WASD}: move/strafe; \texttt{Shift}: run; \texttt{Q/E}: turn L/R; \texttt{J/LB}: fire; \texttt{K/RB}: next weapon; \texttt{L}: prev weapon & \texttt{move(9) $\times$ turn(9) $\times$ attack(5) = 405} \\
\emph{Genshin Impact} & Camera & \texttt{WASD}: move; \texttt{Space}: jump; \texttt{Shift}: dodge dash; \texttt{J}: normal attack; \texttt{LB}: charge attack; \texttt{K/RB}: elemental skill; \texttt{L}: burst; \texttt{Q/E}: swap character; \texttt{R}: lock-on & \texttt{move(9) $\times$ action(12) $\times$ view(9) = 972} \\
\emph{Hexen II} & Camera & \texttt{WASD}: move/strafe; \texttt{Shift}: run; \texttt{Space}: jump; \texttt{L}: crouch; \texttt{Q/E}: turn L/R; \texttt{J/LB}: attack; \texttt{K/RB}: next weapon; \texttt{R}: prev weapon & \texttt{move(9) $\times$ turn(9) $\times$ pitch(3) $\times$ attack(4) = 972} \\
\emph{Hollow Knight} & None & \texttt{A/D}: move L/R; \texttt{W/S}: up/down slash aim; \texttt{Space}: jump; \texttt{Shift}: dash; \texttt{J/LB}: nail slash; \texttt{K}: focus heal & \texttt{move(5) $\times$ jump(9) $\times$ action(22) = 990} \\
\emph{Binding of Isaac} & None & \texttt{WASD}: 8-way move; \texttt{J/K/L/Shift}: fire tears L/D/R/U; \texttt{E}: place bomb & \texttt{move(9) $\times$ fire(5) $\times$ bomb(2) = 90} \\
\emph{Kirby} & None & \texttt{A/D}: walk/fly L/R; \texttt{W}: float up; \texttt{S}: duck; \texttt{Space}: jump; \texttt{J/LB}: inhale / spit star & \texttt{mouth(10) $\times$ walk(9) $\times$ vertical(11) = 990} \\
\emph{KOF '97} & None & \texttt{A/D}: move/block; \texttt{W/Space}: jump; \texttt{S}: crouch; \texttt{J/K}: light punch/kick; \texttt{L/Q}: heavy punch/kick; \texttt{Shift}: roll; \texttt{E}: MAX mode; \texttt{R}: super helper & \texttt{direction(28) $\times$ button(23) = 644} \\
\emph{Super Mario 64} & Camera & \texttt{WASD}: move; \texttt{Space}: jump; \texttt{Shift/RB}: crouch/slide/pound; \texttt{J/LB}: punch/dive/grab; \texttt{K}: slow walk; \texttt{Q/E}: rotate camera 45$^\circ$; \texttt{R}: center camera; \texttt{L}: zoom & \texttt{move(90) $\times$ action(11) = 990} \\
\emph{Mega Man} & None & \texttt{A/D}: move L/R; \texttt{Space}: variable jump; \texttt{J/LB}: buster fire; \texttt{Q/E}: prev/next weapon & \texttt{move(7) $\times$ jump(15) $\times$ fire(4) = 420} \\
\emph{Minecraft} & Camera & \texttt{WASD}: move/strafe; \texttt{Space}: jump; \texttt{Shift}: sprint; \texttt{J/LB}: sword swing; \texttt{RB}: draw/release bow & \texttt{move(6) $\times$ action(6) $\times$ aim(27) = 972} \\
\emph{NFS: Most Wanted} & None & \texttt{W/S}: throttle/brake; \texttt{A/D}: steer L/R; \texttt{Space}: handbrake; \texttt{Shift}: nitrous; \texttt{J}: slow-mo; \texttt{K/L}: shift up/down; \texttt{R}: reset car & \texttt{pedal(4) $\times$ steer(13) $\times$ nitrous(2) $\times$ press(6) = 624} \\
\emph{Quake II} & Camera & \texttt{WASD}: move/strafe; \texttt{Shift}: run; \texttt{Space}: jump; \texttt{L}: crouch; \texttt{Q/E}: turn L/R; \texttt{J/LB}: fire; \texttt{K/RB}: next weapon; \texttt{R}: prev weapon & \texttt{move(9) $\times$ turn(7) $\times$ pitch(3) $\times$ action(5) = 945} \\
\emph{Sekiro} & Camera & \texttt{WASD}: move; \texttt{Space}: jump; \texttt{Shift}: dash/dodge; \texttt{J/LB}: slash/thrust; \texttt{K/RB}: guard/deflect; \texttt{Q}: heal gourd; \texttt{E}: lock-on; \texttt{R}: revive & \texttt{move(12) $\times$ act(73) = 876} \\
\emph{Soul Knight} & Pointer & \texttt{WASD}: 8-way move; \texttt{J/LB}: attack; \texttt{K/RB}: class skill; \texttt{Q}: switch weapon & \texttt{move(9) $\times$ aim(9) $\times$ act(4) = 324} \\
\emph{Terraria} & Pointer & \texttt{A/D}: move L/R; \texttt{Space}: jump; \texttt{S}: drop through platform; \texttt{J/LB}: use weapon; \texttt{Q}: swap sword/bow; \texttt{E}: bow shot at cursor; \texttt{RB}: direct sword swing & \texttt{move(3) $\times$ vert(3) $\times$ attack(19) = 171} \\
\emph{Tomb Raider} & Camera & \texttt{W/S}: forward/back-hop; \texttt{A/D}: tank turn; \texttt{Q/E}: sidestep; \texttt{Space}: jump; \texttt{Shift}: walk; \texttt{J/LB}: fire; \texttt{K/RB}: draw guns; \texttt{L}: 180$^\circ$ roll; \texttt{R}: medipack & \texttt{move(5) $\times$ turn(15) $\times$ act(11) = 825} \\
\emph{Touhou} & None & \texttt{WASD}: 8-way move; \texttt{Shift}: focus slow mode; \texttt{J/LB}: fire; \texttt{K/RB}: bomb; \texttt{Q/E}: switch shot type & \texttt{move(9) $\times$ action(6) = 54} \\
\emph{Zelda: OoT} & Camera & \texttt{WASD}: move; \texttt{Space}: roll/jump; \texttt{Shift}: lock-on; \texttt{J/LB}: sword slash; \texttt{K/RB}: shield block; \texttt{L}: boss item; \texttt{Q/E}: turn camera 45$^\circ$; \texttt{R}: reset camera & \texttt{lock(2) $\times$ move(24) $\times$ act(20) = 960} \\
\bottomrule
\end{tabular}
\end{adjustbox}
\end{table*}

\FloatBarrier
\subsection{Scripted Controller Design}
\label{sec:app-scripted}

\noindent\textbf{Source aware controller synthesis.}
Because \gamedojo{} provides full access to each game's source code, the coding agent writes a dedicated behavior tree controller for every boss encounter.
The agent reads the boss state machine, attack timers, projectile positions, player cooldowns, and the game's legal CUA action space, then implements a deterministic policy hook (exposed through \texttt{BB.GAME.scripted}, \texttt{Module.scripted}, or \texttt{mc\_scripted.scripted}).
Each controller observes live game state strictly in read-only mode, checked before and after every call by verifying that the engine state hash does not change.
To keep demonstrations realistic and aligned with learned agents, every scripted controller runs at the same $200\text{ ms}$ decision step (introducing a natural $200\text{ ms}$ reaction delay) and outputs only legal CUA action packets through the public Gym \texttt{step} function.
In addition, we verify each game's Semantic action space by snapping scripted CUA packets to the nearest Semantic action and confirming that bosses remain winnable.

\noindent\textbf{Difficulty calibration from scripted controllers.}
We use the verified scripted controllers as a consistent skill anchor to calibrate three difficulty levels per encounter.
For each boss, the coding agent tunes numeric difficulty parameters such as boss health, movement speed, attack cooldown, projectile speed, and attack damage, then runs multi seed scripted evaluations until average clear times fall cleanly into three target bands:
\textbf{Easy} ($20\text{--}30\text{ s}$), \textbf{Medium} ($45\text{--}90\text{ s}$), and \textbf{Hard} ($150\text{--}240\text{ s}$).
This procedure ensures that every difficulty tier reflects a smooth progression in combat length and execution pressure while remaining reliably solvable by a strong policy across random seeds.

\noindent\textbf{Two stage trajectory collection.}
We collect \gamedojo{}-Data from the scripted controllers using a two stage parallel pipeline across all 120 bosses and all three difficulty tiers:
\begin{enumerate}
  \item \emph{Headless rollout and filtering}: Parallel workers run the scripted controllers in headless symbolic mode (\texttt{time\_mode=paused}) across distinct engine seeds to record \texttt{actions.jsonl} and \texttt{replay.json}. Only winning episodes that pass exact replay state-hash matching and basic engagement checks (movement ratio $\ge 0.05$, at least $2$ distinct key combinations, and at least $10$ visited arena positions) are kept.
  \item \emph{Replay rendering and dataset assembly}: Accepted action traces are replayed tick-by-tick through the visual renderer to capture pre-action $854\times480$ RGB frames at $60\text{ fps}$, encoded with H.264 (\texttt{libx264}, CRF 18, \texttt{veryfast}, \texttt{yuv420p}). Each video passes strict decode checks (exact frame count matching tick count, SHA-256 verification, and zero corrupt or all-black frames). We remove duplicate action sequences by action-trace hash and partition episodes by \texttt{(boss, seed)} with zero cross-split overlap.
\end{enumerate}
The resulting \gamedojo{}-Data corpus contains $44{,}368$ winning episodes totaling $604.6\text{ h}$: $39{,}937$ training episodes ($543.9\text{ h}$), $2{,}193$ validation episodes ($29.7\text{ h}$), and $2{,}238$ test episodes ($31.0\text{ h}$).

\section{Training Details}
\label{sec:app-training}


\subsection{\gamedojo{}-4B Training Details}
\label{sec:app-sft}

\noindent\textbf{Training datasets.}
For \gamedojo{}-4B (\texttt{S1}), we draw a balanced subset from the training split of \gamedojo{}-Data by sampling complete winning episodes until every boss reaches at least $40$ minutes at each of the three difficulty tiers ($120\times 3\times 40\text{ min} = 240\text{ h}$ nominal, yielding $17{,}859$ complete episodes and $244.9\text{ h}$ total).
To study cross-game transfer (Figure~\gamedojoMainRef{fig:cross-game-transfer}{6} in the main paper) and own-boss learning curves (Figure~\gamedojoMainRef{fig:boss-joint-checkpoints}{4}), we extract matched single-boss ($2\text{ h}$ each), six-boss ($12\text{ h}$ each), and twelve-boss ($24\text{ h}$) subsets from the exact same balanced pool using identical episode selection rules.
Table~\ref{tab:sft-runs} summarizes the dataset statistics and compute budgets for \gamedojo{}-4B and every SFT variant.

\begin{table*}[!htbp]
\centering
\scriptsize
\setlength{\tabcolsep}{2pt}
\renewcommand{\arraystretch}{1.14}
\caption{\textbf{\gamedojo{}-4B and SFT Variant Training Summary.} All runs fine-tune Qwen3.5-4B for 1 epoch with AdamW, peak learning rate $1\times10^{-5}$ (cosine decay to $1\times10^{-6}$), and $4.0\text{ s}$ (up to 20 step) causal windows. In the main paper, \texttt{S1} supports Table~\gamedojoMainRef{tab:difficulty}{4} and \texttt{S1}--\texttt{S6} support Figure~\gamedojoMainRef{fig:cross-game-transfer}{6}; \texttt{S5}, \texttt{S6}, \texttt{S2}, \texttt{S3}, and the 10 additional single-boss 2\,h runs (\texttt{S7}--\texttt{S16}) support Figure~\gamedojoMainRef{fig:boss-joint-checkpoints}{4}.}
\label{tab:sft-runs}
\begin{adjustbox}{max width=\linewidth}
\begin{tabular}{@{}l>{\raggedright\arraybackslash}p{85pt}rrrrrrrl@{}}
\toprule
\textbf{ID} & \textbf{Training Bosses} & \textbf{Episodes} & \textbf{Hours} & \textbf{4\,s Windows} & \textbf{CUA Actions} & \textbf{Batch} & \textbf{Steps} & \textbf{GPUs} & \textbf{A800 GPU-h} \\
\midrule
\texttt{S1} (\gamedojo{}-4B) & All 120 bosses (20 games) & 17,859 & 244.9 & 229,423 & 4,415,457 & 512 & 449 & 48 & 344.2 \\
\texttt{S2} & Terraria (all 6 bosses) & 933 & 12.2 & 11,488 & 220,189 & 64 & 180 & 16 & 16.0 \\
\texttt{S3} & Doom (all 6 bosses) & 877 & 12.3 & 11,455 & 221,089 & 64 & 179 & 16 & 16.2 \\
\texttt{S4} & Terraria + Doom (12 bosses) & 1,810 & 24.5 & 22,943 & 441,278 & 64 & 359 & 16 & 33.2 \\
\texttt{S5} & Terraria: Queen Bee (1 boss) & 163 & 2.0 & 1,902 & 36,527 & 64 & 30 & 16 & 4.5 \\
\texttt{S6} & Doom: Cyberdemon (1 boss) & 152 & 2.1 & 1,925 & 37,090 & 64 & 31 & 16 & 4.5 \\
\midrule
\texttt{S7}--\texttt{S11} (5 runs) & Other 5 Terraria bosses (each 1 boss) & $\sim$155/run & $\sim$2.0/run & 1,906--1,931 & 36,390--36,949 & 64 & 30--31 & 16 & $\sim$4.5/run \\
\texttt{S12}--\texttt{S16} (5 runs) & Other 5 Doom bosses (each 1 boss) & $\sim$150/run & $\sim$2.0/run & 1,882--1,932 & 36,520--37,210 & 64 & 30--31 & 16 & $\sim$4.5/run \\
\bottomrule
\end{tabular}
\end{adjustbox}
\end{table*}

\noindent\textbf{Training hyperparameters and compute budget.}
Every SFT run starts from the pretrained \texttt{Qwen3.5-4B} vision-language backbone (24 linear-attention \texttt{GatedDeltaNet} layers interleaved with 8 full-attention layers).
Each episode is segmented into non overlapping $4.0\text{ s}$ causal multi-turn windows of up to 20 consecutive $(\text{frame}, \text{CUA action})$ turns paired with the exact English system prompt used during evaluation.
All frames are passed at native $854\times480$ resolution (\texttt{min\_pixels=65536}, \texttt{max\_pixels=414720}).
Because scripted trajectories contain no chain-of-thought text, every assistant turn begins with a closed empty thinking prefix (\texttt{<think>\textbackslash n\textbackslash n</think>}) immediately followed by the CUA action string and end-of-turn token, matching our non-thinking inference template (\texttt{enable\_thinking=False}).
Loss is computed exclusively on the assistant action tokens and end-of-turn tokens (system prompts, images, user headers, and empty thinking tags are masked out) using token-averaged FP32 cross-entropy over each global batch.

We train all models for 1 epoch with AdamW ($\beta_1=0.9$, $\beta_2=0.999$, weight decay $0.01$, gradient norm clipping $1.0$), a peak learning rate of $1\times10^{-5}$, linear warmup over the first $\sim5\%$ of updates (22 steps for \texttt{S1}, 2--18 steps for variants), and cosine decay to $1\times10^{-6}$.
To keep throughput high without cross-sample attention leakage, multiple $4.0\text{ s}$ windows are packed up to $12{,}000$ tokens per GPU with strict boundary isolated attention masks and per segment position IDs (each window spans $\sim2{,}500\text{--}10{,}500$ tokens).
Training runs under PyTorch FSDP2 (\texttt{fsdp\_size=8}, \texttt{bfloat16} activations and compute, FP32 master weights and gradient reductions) on NVIDIA A800-SXM4-80GB GPUs.
Total SFT compute is $344.2$ GPU hours for \gamedojo{}-4B (\texttt{S1}, 48 GPUs), $74.4$ GPU hours for the five transfer variants (\texttt{S2}--\texttt{S6}, 16 GPUs each), and $\sim45.0$ GPU hours for the ten additional single-boss curve runs (\texttt{S7}--\texttt{S16}, 16 GPUs each).

\noindent\textbf{Complete set of SFT experiments.}
Our SFT experiments comprise three groups:
\begin{enumerate}
  \item \emph{All boss generalist training} (\texttt{S1}, \gamedojo{}-4B): trained on 244.9\,h across all 120 bosses and evaluated on all 120 bosses across Easy, Medium, and Hard in both time-paused and real-time CUA modes (Table~\gamedojoMainRef{tab:difficulty}{4} of the main paper).
  \item \emph{Cross game and within game transfer} (\texttt{S2}--\texttt{S6}, compared against \texttt{S1} and base Qwen3.5-4B): five source configurations spanning 1, 6, and 12 bosses from \emph{Terraria} and \emph{Doom}, evaluated on all 120 bosses grouped into Terraria, Doom, other 2D games, other 3D games, and all 120 bosses (Figure~\gamedojoMainRef{fig:cross-game-transfer}{6} of the main paper).
  \item \emph{Own-boss SFT learning curves} (\texttt{S5}, \texttt{S6}, \texttt{S7}--\texttt{S16} for single-boss 2\,h runs, plus \texttt{S2} and \texttt{S3} for 6-boss 12\,h joint runs): checkpoints saved at 20\%, 40\%, 60\%, 80\%, and 100\% of 1 epoch to trace own-boss progress against PPO in Figure~\gamedojoMainRef{fig:boss-joint-checkpoints}{4} of the main paper.
\end{enumerate}

\gamedojoSupplementFloatBarrier
\subsection{PPO Training Details}
\label{sec:app-ppo}

\noindent\textbf{Training rollouts and per run summary.}
We train 14 visual PPO policies from scratch across \emph{Terraria} (2D) and \emph{Doom} (3D): six single-boss specialists (\texttt{P1}--\texttt{P6} in \emph{Terraria}, \texttt{P8}--\texttt{P13} in \emph{Doom}) and one six-boss joint policy per game (\texttt{P7} in \emph{Terraria}, \texttt{P14} in \emph{Doom}).
During training rollouts, the policy samples an action from the categorical distribution of each Semantic action head.
Each single-boss run uses $100{,}000$ environment steps ($33$ PPO updates, $12{,}500$ Adam minibatch steps), while each six-boss joint run uses $600{,}000$ environment steps ($196$ PPO updates, $75{,}000$ Adam minibatch steps), matching the exact exposure of $100{,}000$ steps per boss.
Table~\ref{tab:ppo-runs} lists every PPO run, its action space, rollout steps, completed training episodes, wall clock duration, and GPU hours.

\begin{table*}[!htbp]
\centering
\scriptsize
\setlength{\tabcolsep}{4.5pt}
\renewcommand{\arraystretch}{1.12}
\caption{\textbf{All 14 PPO Training Runs.} Every run trains a shared Nature-CNN policy-value network from scratch on Easy difficulty using the game's Semantic action space ($0.2\text{ s}$ decision interval) and signed step-damage reward on a single NVIDIA A800-SXM4-80GB GPU.}
\label{tab:ppo-runs}
\begin{adjustbox}{max width=\linewidth}
\begin{tabular}{@{}lllrrrrrr@{}}
\toprule
\textbf{ID} & \textbf{Game} & \textbf{Training Boss(es)} & \textbf{Sem.\ Size} & \textbf{Env Steps} & \textbf{Updates} & \textbf{Episodes} & \textbf{Minutes} & \textbf{A800 GPU-h} \\
\midrule
\texttt{P1} & \emph{Terraria} & King Slime & 171 & 100,000 & 33 & 1,085 & 15.6 & 0.26 \\
\texttt{P2} & \emph{Terraria} & Eye of Cthulhu & 171 & 100,000 & 33 & 306 & 15.6 & 0.26 \\
\texttt{P3} & \emph{Terraria} & Skeletron & 171 & 100,000 & 33 & 436 & 15.6 & 0.26 \\
\texttt{P4} & \emph{Terraria} & Eater of Worlds & 171 & 100,000 & 33 & 361 & 15.6 & 0.26 \\
\texttt{P5} & \emph{Terraria} & Queen Bee & 171 & 100,000 & 33 & 483 & 15.6 & 0.26 \\
\texttt{P6} & \emph{Terraria} & Brain of Cthulhu & 171 & 100,000 & 33 & 365 & 15.6 & 0.26 \\
\texttt{P7} & \emph{Terraria} & All 6 Terraria bosses (joint) & 171 & 600,000 & 196 & 3,603 & 93.6 & 1.56 \\
\midrule
\texttt{P8} & \emph{Doom} & Cyberdemon & 405 & 100,000 & 33 & 1,524 & 15.8 & 0.26 \\
\texttt{P9} & \emph{Doom} & Barons of Hell & 405 & 100,000 & 33 & 433 & 15.6 & 0.26 \\
\texttt{P10} & \emph{Doom} & Spider Mastermind & 405 & 100,000 & 33 & 687 & 15.7 & 0.26 \\
\texttt{P11} & \emph{Doom} & Archvile & 405 & 100,000 & 33 & 749 & 15.7 & 0.26 \\
\texttt{P12} & \emph{Doom} & Mancubus & 405 & 100,000 & 33 & 344 & 15.7 & 0.26 \\
\texttt{P13} & \emph{Doom} & Revenant & 405 & 100,000 & 33 & 277 & 15.6 & 0.26 \\
\texttt{P14} & \emph{Doom} & All 6 Doom bosses (joint) & 405 & 600,000 & 196 & 3,970 & 93.7 & 1.56 \\
\midrule
\textbf{Total} & \textbf{2 Games} & \textbf{14 PPO Policies} & -- & \textbf{2,400,000} & \textbf{788} & \textbf{14,623} & \textbf{374.9} & \textbf{6.25} \\
\bottomrule
\end{tabular}
\end{adjustbox}
\end{table*}

\noindent\textbf{Architecture, hyperparameters, and compute budget.}
Following DSLE~\citep{gezgin2026dsle}, each policy uses a shared Nature-style CNN encoder: three convolutional layers (\texttt{Conv2d(1, 32, 8, stride=4)}, \texttt{Conv2d(32, 64, 4, stride=2)}, \texttt{Conv2d(64, 64, 3, stride=1)}) followed by a 512-unit fully connected layer with ReLU activations.
The 512-dimensional feature vector feeds a scalar value head and one independent linear classification head per Semantic action dimension ($3+3+19$ logits in \emph{Terraria}, $1{,}691{,}322$ parameters; $9+9+5$ logits in \emph{Doom}, $1{,}690{,}296$ parameters).
Observations are single frame (\texttt{frame\_stack=1}) $84\times84$ \texttt{uint8} grayscale images converted from RGB via $0.299R + 0.587G + 0.114B$ and adaptive area resizing.
Training runs on Easy difficulty ($d=0.2$ in \emph{Terraria}, $d=0.15$ in \emph{Doom}) with the signed per step normalized boss-damage reward $r_t = (H_t - H_{t+1})/H_{\max}$ and no extra player-damage, time, or terminal bonus terms.

We use 24 parallel environment slots (6 worker processes $\times$ 4 environments per worker), a rollout horizon of $128$ steps ($3{,}072$ transitions per update), $4$ PPO epochs per update, and minibatch size $32$.
Optimization uses Adam with initial learning rate $2.5\times10^{-4}$ linearly annealed to $0$, $\epsilon=10^{-5}$, discount $\gamma=0.99$, GAE $\lambda=0.95$, clipping ratio $0.1$, entropy coefficient $0.01$, value loss coefficient $0.5$, maximum gradient norm $0.5$, no value clipping, and FP32 precision with TF32 disabled.
Each run uses one NVIDIA A800-SXM4-80GB GPU, averaging $106.4\text{--}106.9$ environment steps per second.
A single-boss run finishes in $15.6\text{--}15.8$ minutes ($0.26$ GPU-h) and a six-boss joint run finishes in $93.6\text{--}93.7$ minutes ($1.56$ GPU-h), totaling $6.25$ A800 GPU hours for all 14 models.

\noindent\textbf{Complete set of PPO experiments.}
We conduct three evaluations over the 14 trained policies:
\begin{enumerate}
  \item \emph{Single-boss learning curves} (\texttt{P1}--\texttt{P6}, \texttt{P8}--\texttt{P13}): track episode progress over $100{,}000$ steps on each boss (Figure~\gamedojoMainRef{fig:ppo-boss-learning}{3}).
  \item \emph{Single-boss vs.\ joint training} (\texttt{P7} vs.\ \texttt{P1}--\texttt{P6} in \emph{Terraria}; \texttt{P14} vs.\ \texttt{P8}--\texttt{P13} in \emph{Doom}): compare shared six-boss training against dedicated single-boss training at equal per boss budget (Figure~\gamedojoMainRef{fig:boss-joint-checkpoints}{4} and Section~\gamedojoMainRef{sec:analysis-boss-transfer}{5.2}).
  \item \emph{Zero-shot cross-boss transfer}: freeze each of the 12 single-boss checkpoints and evaluate on all 6 bosses in its own game, yielding 30 unseen source-to-target pairs per game (60 pairs total) compared against randomly initialized PPO policies. Figure~\gamedojoMainRef{fig:boss-transfer-checkpoints}{5} of the main paper shows the fixed-source comparison.
\end{enumerate}

\gamedojoSupplementFloatBarrier
\Needspace{11\baselineskip}
\section{Evaluation Details}
\label{sec:app-eval}


\subsection{Benchmark Evaluation Protocol}
\label{sec:app-protocol}

\noindent\textbf{Time-paused vs.\ real-time modes.}
\gamedojo{} evaluates every environment under two timing protocols that share the exact same $60\text{ Hz}$ physics engine and $200\text{ ms}$ decision step:
\begin{itemize}
  \item \textbf{Time-paused mode (TP)}: The environment renders one $854\times480$ RGB frame and immediately freezes its $60\text{ Hz}$ clock while the agent runs inference. Only after the action arrives does the simulation advance by exactly one $200\text{ ms}$ step ($12$ ticks), after which it freezes again for the next observation. Inference speed therefore has zero effect on episode outcomes.
  \item \textbf{Real-time mode (RT)}: The simulation clock stays coupled to wall clock inference time. If a model call takes $\Delta t$ seconds from receiving the observation to returning an action, the environment first advances by $\lfloor \Delta t / 0.2\text{ s} \rfloor$ neutral no op steps ($200\text{ ms}$ each, with all keys released and zero mouse displacement) before applying the returned action to the resulting state. Latency below $200\text{ ms}$ incurs zero lag steps. If the player dies or the boss is defeated during the no op lag steps, the episode ends on that exact tick and the late action is discarded.
\end{itemize}

\noindent\textbf{Evaluation forms across agent families.}
All three agent families are scored by the same encounter-level win rate (\texttt{WR}, \%) and boss-health progress (\texttt{PG}, \%), averaged first over episodes within each boss and then with equal weight over all evaluated bosses:
\begin{itemize}
  \item \textbf{VLM and LLM agents}: Receive $854\times480$ RGB frames, public control instructions, and a 2 step sliding history of recent frames and actions. Frontier API models are evaluated on all 120 bosses with a $300$ step cap ($60\text{ s}$ game time) in TP mode and a $180\text{ s}$ wall clock budget in RT mode. Self-hosted \texttt{Qwen3.5-4B} and \gamedojo{}-4B checkpoints run $5$ independent seeds per boss ($600$ episodes per difficulty and timing mode) with a $180\text{ s}$ ($900$ step) budget.
  \item \textbf{PPO specialist policies}: Receive only the current $84\times84$ grayscale frame with no text prompt or frame history, outputting Semantic actions at $0.2\text{ s}$ steps up to the $300\text{ s}$ engine limit. Each policy is evaluated for $20$ episodes per boss using test seeds held out from training.
  \item \textbf{Scripted controllers}: State-aware behavior trees running through the same CUA step interface as a privileged upper-bound reference, evaluated over $5$ seeds per boss ($600$ episodes per difficulty) under matched $60\text{ s}$ ($300$ step, TP comparison) and $180\text{ s}$ ($900$ step, RT comparison) game-time budgets.
\end{itemize}

\subsection{VLM Evaluation Details}
\label{sec:app-vlm-eval}

\noindent\textbf{Shared evaluation harness and prompt specification.}
All frontier, open-source, and generalist game models run through a unified OpenAI compatible evaluation harness (\texttt{bossbench\_llm}).
At episode start, the harness builds a deterministic system prompt by concatenating five fixed sections:
(1)~role and $200\text{ ms}$ step timing rule,
(2)~exact output syntax for either CUA (6-substep token string) or Semantic (named JSON dictionary),
(3)~the game overview and the active controls for the interface,
(4)~concise boss mechanics (excluding numeric health values, which must be read from the screen), and
(5)~a short footer with output formatting or observation instructions.
At each decision step, the current $854\times480$ frame is JPEG-encoded (quality 90, base64) and appended to the conversation while retaining a 2 step FIFO window of the two most recent \texttt{(image, action)} pairs.
Invalid actions count as format errors and execute a neutral no-op. The harness retries temporary network or HTTP failures up to three times and excludes those failures from model error counts.
Figure~\ref{fig:prompt-templates} shows the CUA and Semantic system prompts for the King Slime encounter in \emph{Terraria}.

\begin{figure*}[!htbp]
\centering
\scriptsize
\setlength{\fboxsep}{6pt}
\setlength{\fboxrule}{0.5pt}
\begin{minipage}[t]{0.488\textwidth}
\fcolorbox{black!45}{black!3}{%
\begin{minipage}[t]{\dimexpr\linewidth-2\fboxsep-2\fboxrule\relax}
\raggedright
\textbf{CUA System Prompt (Example: \emph{Terraria} / King Slime)}\\[3pt]
\hrule\vspace{3pt}
\texttt{You are an experienced PC game player, proficient in keyboard and mouse operations. You are playing "Pixel Arena: Sword and Bow". Based on the current game screen and the corresponding historical trajectory, plan the next 200ms of actions, consisting of 6 steps. Each step is spaced 33ms apart. Every step lasts 33ms from its start time until the next step begins.}\\[3pt]
\textbf{Output Format}\\
{\fontsize{5.6}{6.8}\selectfont\texttt{<|action\_start|>X Y Z ; k1 k2 ; k3 ; k4 ; k5 ; k6 ; k7<|action\_end|>}}\\
\textbf{Explanation}\\
1. \textbf{Mouse Movement}: Relative integers \texttt{X, Y} in \texttt{[-1000, 1000]} (\texttt{X>0} right, \texttt{Y>0} down) and scroll \texttt{Z} in \texttt{[-5, 5]} (\texttt{Z>0} up).\\
2. \textbf{Key Sequence}: 6 groups separated by \texttt{;}; space-separated keys within each group (at most 4 keys/group). Leave an empty group blank while keeping \texttt{;}. Keys listed in consecutive groups stay held; unlisted keys are released.\\
3. Output only one plain line matching the format, with no line breaks or quotes.\\
\textbf{Key Naming Rules}: Capital letters (\texttt{W}, \texttt{A}), capitalized names (\texttt{Space}, \texttt{Shift}), mouse buttons (\texttt{LB}, \texttt{RB}).\\[3pt]
\textbf{Game}: Side-view 2D arena with two wooden platforms. Empty the boss's health bar within 300 s without losing all health.\\
\textbf{Controls} (only these keys are active): \texttt{A}/\texttt{D} move left/right; \texttt{Space} jump/fly; \texttt{S} drop through platform; \texttt{J}/\texttt{LB} use weapon; \texttt{Q} swap sword/bow; \texttt{E} shoot bow at crosshair; \texttt{RB} swing sword.\\
\textbf{Mouse}: \texttt{X/Y} move crosshair (\texttt{1000} = full screen width/height); bow shoots toward crosshair.\\
\textbf{Boss: King Slime}: Giant gel creature that leaps to crush you, sheds small blobs when hit, and teleports near you. Moves: 1. Leap; 2. Teleport; 3. Gel Spawns.\\
\texttt{Health bars and other status are shown on the screen; read them yourself.}
\end{minipage}}
\end{minipage}
\hfill
\begin{minipage}[t]{0.488\textwidth}
\fcolorbox{black!45}{black!3}{%
\begin{minipage}[t]{\dimexpr\linewidth-2\fboxsep-2\fboxrule\relax}
\raggedright
\textbf{Semantic System Prompt (Example: \emph{Terraria} / King Slime)}\\[3pt]
\hrule\vspace{3pt}
\texttt{You are an experienced PC game player. You are playing "Pixel Arena: Sword and Bow". Based on the current game screen and the corresponding historical trajectory, choose the action for the next 200ms.}\\[3pt]
\textbf{Output Format}\\
Output one JSON object on a single line that picks one option for each action group, for example:\\
{\fontsize{6.2}{7.4}\selectfont\texttt{\{"move": "left", "vert": "jump", "attack": "shoot 135"\}}}\\
- All groups act at the same time for the next 200ms.\\
- Any omitted group takes its first option (\texttt{none} = do nothing).\\
- Use option names exactly as listed; do not output indices or any other text.\\[3pt]
\textbf{Action Groups}\\
- \texttt{move} (horizontal): \texttt{none} (stop); \texttt{left} (move left); \texttt{right} (move right).\\
- \texttt{vert} (vertical): \texttt{none}; \texttt{jump} (jump/fly); \texttt{drop} (drop through platform).\\
- \texttt{attack} (weapon): \texttt{none}; \texttt{shoot <angle>} (bow arrow at 0--359$^\circ$, 0$^\circ$=right, 90$^\circ$=up, rounded to 22.5$^\circ$); \texttt{sword\_left}; \texttt{sword\_right}.\\[3pt]
\textbf{Game}: Side-view 2D arena with two wooden platforms. Empty the boss's health bar within 300 s without losing all health.\\
\textbf{Mechanics}: Flight lasts $\sim$1.5 s and refills on landing; bow fires every 0.33 s and sword swings every 0.38 s; both destroy boss projectiles; brief invulnerability after taking a hit.\\
\textbf{Boss: King Slime}: Giant gel creature that leaps to crush you, sheds small blobs when hit, and teleports near you. Moves: 1. Leap; 2. Teleport; 3. Gel Spawns.\\
\texttt{Health bars and other status are shown on the screen; read them yourself.}
\end{minipage}}
\end{minipage}
\caption{\textbf{System Prompt Templates for CUA (left) and Semantic (right) Interfaces.} Illustrated with \emph{Terraria} (King Slime). The examples show the CUA key sequence and the Semantic action groups. Both prompts state the $300\text{ s}$ arena limit. The benchmark evaluation budgets are given in Section~\ref{sec:app-protocol}.}
\label{fig:prompt-templates}
\end{figure*}

\noindent\textbf{Closed source models.}
Table~\ref{tab:eval-closed} lists the reasoning effort setting, average output tokens per decision step (including reasoning tokens when produced), and average per step latency for all six closed-source API models evaluated in Table~\gamedojoMainRef{tab:frontier-main}{3} of the main paper.
per step token statistics are measured across the completed time-paused CUA benchmark runs (21,622--27,496 decisions per model; 14,421 decisions on Hard for Claude-Haiku-5.5).
Where per step API wall clock latency was not logged in the aggregated run summary, we mark the entry with \texttt{-}.

\begin{table}[!htbp]
\centering
\scriptsize
\setlength{\tabcolsep}{2.5pt}
\renewcommand{\arraystretch}{1.14}
\caption{\textbf{Closed-Source Model Evaluation Settings.} Average output tokens per step include emitted reasoning tokens measured on time-paused CUA runs. \texttt{-} denotes fields not recorded in the aggregated API logs.}
\label{tab:eval-closed}
\begin{tabular}{@{}llcc@{}}
\toprule
\textbf{Model} & \textbf{Reasoning Effort} & \shortstack[c]{\textbf{Avg.\ Out}\\\textbf{Tok/Step}} & \shortstack[c]{\textbf{Avg.}\\\textbf{Latency}} \\
\midrule
GPT-6.1-Sol & \texttt{low} & 42.9 & - \\
GPT-6-Luna & \texttt{low} & 227.9 & - \\
Claude-Opus-5.5 & \texttt{low} & 67.9 & - \\
Claude-Haiku-5.5 & \texttt{low} & 183.4 & - \\
Claude-Haiku-4.5 & \texttt{low} (thinking on) & 478.3 & - \\
Gemini-3.8-Flash & \texttt{low} & 75.6 & - \\
\bottomrule
\end{tabular}
\end{table}

\noindent\textbf{Open-source and generalist game models.}
Table~\ref{tab:eval-open} summarizes reasoning mode, average output tokens per step, average latency per step, and deployment settings for all open-source VLMs and generalist game models.
For our self-hosted \texttt{Qwen3.5-4B} and \gamedojo{}-4B evaluations in Table~\gamedojoMainRef{tab:difficulty}{4} of the main paper, we serve exported \texttt{bfloat16} checkpoints with vLLM on single NVIDIA A800-SXM4-80GB GPUs (\texttt{tensor\_parallel\_size=1}, \texttt{max\_model\_len=16384}, prefix caching enabled, \texttt{limit\_mm\_per\_prompt=\{"image":3,"video":0\}}).
Inference uses the aligned non-thinking chat template (\texttt{enable\_thinking=False}), prefilling \texttt{<think>\textbackslash n\textbackslash n</think>} so the model emits CUA tokens directly (\texttt{temperature=1.0}, \texttt{top\_p=1.0}, \texttt{top\_k=-1}, \texttt{max\_tokens=256}, \texttt{skip\_special\_tokens=}\allowbreak\texttt{False}), with streaming abort as soon as \texttt{<|action\_}\allowbreak\texttt{end|>} is generated.
Under 8 concurrent environment workers sharing one A800 replica, \gamedojo{}-4B averages $32.8\text{--}33.5$ output tokens per step, $0.72\text{--}0.77\text{ s}$ mean round trip latency in RT mode ($0.88\text{--}0.92\text{ s}$ in TP mode), and $99.96\%\text{--}99.98\%$ action format validity across all difficulties.

\begin{table*}[!htbp]
\centering
\scriptsize
\setlength{\tabcolsep}{2.5pt}
\renewcommand{\arraystretch}{1.16}

\caption{\textbf{Open Source and Generalist Game Model Evaluation Settings.} Average output tokens per step include reasoning tokens where thinking is enabled. \texttt{stop tag} indicates streaming termination upon emitting \texttt{<|action\_end|>}}
\label{tab:eval-open}
\begin{adjustbox}{max width=\linewidth}
\begin{tabular*}{\textwidth}{@{\extracolsep{\fill}}lcccl>{\raggedright\arraybackslash}p{0.235\textwidth}@{}}
\toprule
\textbf{Model} & \textbf{Reasoning} & \textbf{Out Tok/Step} & \textbf{Latency (s)} & \textbf{Serving \& GPU} & \textbf{Decoding Setup} \\
\midrule
\rowcolor{black!8}
\multicolumn{6}{@{}l}{\textit{Open Source API Models}} \\
DeepSeek-V4.1-Flash & On & 881.0 & - & Hosted API & API default \\
GLM-5.3-Flash & On & 1941.1 & - & Hosted API & API default \\
MiniMax-M3 & On & 1482.7 & - & Hosted API & API default \\
\midrule
\rowcolor{black!8}
\multicolumn{6}{@{}l}{\textit{Self-Hosted Open Source Models}} \\
Qwen3.5-35B-A3B & On & 1357.1 & - & Local vLLM (BF16) & 2,048-tok thinking budget \\
Qwen3.5-4B (Main Table~\gamedojoMainRef{tab:frontier-main}{3}) & On & 1819.7 & - & Local vLLM (BF16) & 2,048-tok thinking budget \\
Qwen3.5-4B (Main Table~\gamedojoMainRef{tab:difficulty}{4}) & Off & 46.6 & 0.90 & Local vLLM (BF16) & None-thinking \\
\gamedojo{}-4B (S1) & Off & 32.8 & 0.74 &  Local vLLM (BF16) & Non-thinking \\
\midrule
\rowcolor{black!8}
\multicolumn{6}{@{}l}{\textit{Generalist Game Models (CUA Only)}} \\
GameScaling-9B-S3 & Off & - & - & Local vLLM (BF16) & Non-thinking \\
Gamer-35B-A3B & Off & - & - & Local vLLM (BF16) & Non-thinking \\
\bottomrule
\end{tabular*}
\end{adjustbox}
\end{table*}

\FloatBarrier
\subsection{PPO Evaluation Details}
\label{sec:app-ppo-eval}

\noindent\textbf{Average inference latency.}
Because the specialist PPO policy is a compact 1.69M-parameter Nature CNN operating on $1\times84\times84$ grayscale inputs, policy inference time is very small compared with the $200\text{ ms}$ decision step.
In our NVIDIA A800-SXM4-80GB throughput benchmark (8 parallel browser actor processes $\times$ 8 environment slots = 64 concurrent environments), batched policy forward passes and action selection take $7.14\text{ s}$ across $43{,}465$ decisions in \emph{Terraria} ($0.16\text{ ms}$ per decision, $776.4$ end to end steps/s) and $12.87\text{ s}$ across $70{,}544$ decisions in \emph{Doom} ($0.18\text{ ms}$ per decision, $575.0$ end to end steps/s).
Even single frame forward passes execute in well under $1\text{ ms}$ on GPU, which is more than two orders of magnitude below the $200\text{ ms}$ decision interval and therefore introduces zero real-time lag steps.

\noindent\textbf{Action selection.}
All PPO evaluations use greedy \texttt{argmax}.
At each decision step, the policy selects the option with the highest logit in each Semantic action head and combines the selected options into one action.